\documentclass[11pt]{article}
\usepackage{acl}

\usepackage{times}
\usepackage{latexsym}

\usepackage[T1]{fontenc}
\usepackage[utf8]{inputenc}

\usepackage{microtype}
\usepackage{inconsolata}

\usepackage{graphicx}
\usepackage{amsmath}
\usepackage{rotating}
\usepackage{booktabs}
\usepackage{tabularx}
\usepackage{multirow}
\usepackage{tcolorbox}
\usepackage{placeins}
\usepackage{hyperref}
\usepackage{listings}
\usepackage{xcolor}

\usepackage[normalem]{ulem}
\useunder{\uline}{\ul}{}

\definecolor{fc}{RGB}{173, 76, 39}
\definecolor{ngo}{RGB}{46, 120, 160}

\newtcolorbox{ngobox}[1][]{
  colframe=ngo,
  colback=white,
  boxrule=1pt,
  arc=5pt,
  left=10pt,
  right=10pt,
  top=10pt,
  bottom=10pt,
  #1
}

\newtcolorbox{fcbox}[1][]{
  colframe=fc,
  colback=white,
  boxrule=1pt,
  arc=5pt,
  left=10pt,
  right=10pt,
  top=10pt,
  bottom=10pt,
  #1
}

\title{CATCH-ME if you RAG:\\ a dataset of Contextually Annotated multi-Turn Counterspeech\\ against Hate and Misinformation Exchanges}

\author{
Helena Bonaldi\textsuperscript{1} \quad
Genoveffa Martone\textsuperscript{1,2} \quad
Marco Guerini\textsuperscript{1} \\
\textsuperscript{1}Fondazione Bruno Kessler, Italy,
\textsuperscript{2}Università Cattolica del Sacro Cuore, Italy, \\
\texttt{\{hbonaldi, gmartone, guerini\}@fbk.eu}
}

\begin{document}
\maketitle
\begin{abstract}
Online hate speech and misinformation frequently overlap, yet NLP research has mainly treated them in isolation. While LLMs represent a scalable solution for assisting humans in the generation of counterspeech for both threats, zero-shot models frequently generate repetitive and vague responses, underscoring the need for high-quality examples to steer model generation. However, existing counterspeech datasets against the overlap of hate and misinformation are scarce and limited to single-turn English dialogues, while real-life interactions span across multiple turns and languages. To bridge this gap, we introduce the first large-scale, expert-curated, multilingual dataset of dialogues tackling the intersection of hate and misinformation. To ensure factual grounding, the dialogues are also anchored in verified external knowledge (i.e., fact-checking articles and NGO reports) and include document- and chunk-level span annotations, making it directly applicable for RAG systems. Covering five languages and targeting hate directed at seven marginalized groups, this novel resource enables the training and evaluation of more persuasive, factually grounded counterspeech models.

\textit{\textbf{Warning}: This work contains unobfuscated examples that some readers may find offensive.}

\end{abstract}

\begin{figure}[t!]
    \centering
    \includegraphics[width=0.9\columnwidth]{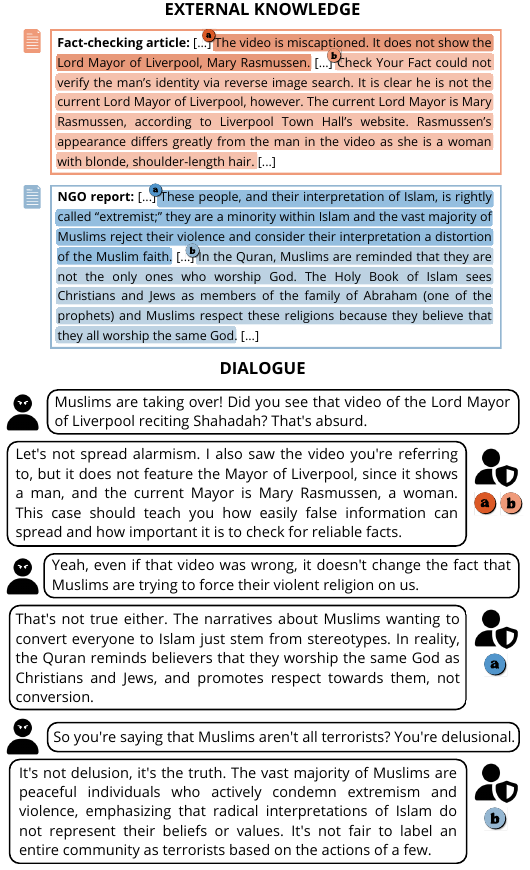}
    \caption{An example of a collected dialogue, where each CS is supported by 
    the external knowledge.}
    \label{fig:dialogue_ex}
\end{figure}
\section{Introduction}
While online hate and misinformation are traditionally treated as separate NLP problems \cite{chung2021empowering, zeng-gao-2024-justilm}, they frequently interact in real-world settings \cite{cazzamatta2025global}. Hate speech co-occurs more with misinformation than with factual content \cite{hameleers2022civilized}, and misinformation often exacerbates hate speech, amplifying its reach and potential for harm \cite{wardle2024conceptual}. In this context, counterspeech (CS) has emerged as an alternative to combat these phenomena without resorting to censorship \cite{myers2018censored}. While CS against hate uses empathy and cogent, fact-based arguments \cite{schieb2016governing}, CS against misinformation adopts a non-partisan, non-emotive style 
\cite{guo2022survey}. Conversely, CS tackling their intersection must jointly correct misinformation with facts and use empathy to challenge negative stereotypes \cite{martone2026assisted}. Although NGOs and fact-checkers manually produce CS \cite{conan-2019, wintersieck2017debating}, this process requires high cognitive effort and psychological resilience, creating a bottleneck for moderators \cite{mun2024counterspeakers}.
Consequently, interest has surged in leveraging Large Language Models (LLMs) to assist them in writing CS \cite{bonaldi-etal-2024-nlp}. However, state-of-the-art LLMs in zero-shot settings often produce repetitive, vague responses that denounce rather than constructively engage with harmful content \cite{munbeyond}. This underscores the need for high-quality examples to steer generation. Yet, existing CS corpora isolate these issues, focusing either on toxic content or objective veracity \cite{bonaldi-etal-2024-nlp, guo2022survey}. The only dataset addressing both is restricted to single-turn English dialogues \cite{martone2026assisted}.
To bridge this gap, we collaborate with 23 experts in CS writing and employ four human-machine collaboration strategies to create \textbf{CATCH-ME}: a dataset of \textbf{C}ontextually \textbf{A}nnotated multi-\textbf{T}urn \textbf{C}ounterspeech against \textbf{H}ate and \textbf{M}isinformation \textbf{E}xchanges. CATCH-ME includes fictitious multi-turn interactions between a person spreading hate and misinformation, and a counterspeaker addressing both.
Responses are grounded in external verified knowledge, including fact-checking articles and NGO reports annotated at both chunk and document level, making the dataset well suited for Retrieval-Augmented Generation (RAG) research. An example of a collected dialogue with annotated external knowledge is shown in Figure \ref{fig:dialogue_ex}.
The dataset covers five languages (English, Italian, Maltese, Polish, and Spanish) and targets hostility toward seven marginalized groups: Muslims, Jewish people, people of color, women, LGBTQIA+ individuals, migrants, and people with disabilities.
We describe our collection and annotation methodologies across the four collaboration strategies, and use our dataset to provide a first benchmark on retrieval and generation tasks for fact-based CS. The final dataset contains 2015 dialogues and 12,298 turns in total, 6,149 of which are CS turns grounded in external knowledge.\footnote{The dataset is available at \url{https://github.com/LanD-FBK/counterspeech_against_hate_and_misinfo}.}

\section{Related Work}
\paragraph{CS data collection}
The techniques employed to collect CS resources include \textit{crawling} 
them from online platforms \citep{mathew2019thou} or fact-checking websites \citep{russo2023benchmarking};
\textit{crowdsourcing} 
them from non-expert annotators \cite{furman-etal-2023-high, he2023reinforcement} 
or \textit{nichesourcing} them from domain experts, such as NGO operators \cite{chung2021towards} and professional fact-checkers \cite{russo2025euroverdict}. CS can also be obtained with \textit{fully automated} methods
\cite{vallecillo2023automatic, stammbach2020fever}: their use, however, remains limited due to concerns about the factual accuracy and potential harmfulness of their outputs.
Finally, \textit{human-in-the-loop} (HITL) approaches combine automated generation with human post-editing to balance scalability and quality \cite{tekiroglu2020generating, fanton2021human, russo2023countering}. In this context, \citet{martone2026assisted} introduces the only HITL dataset addressing the intersection of hate and misinformation, which
is limited to single-turn interactions. 
The only available conversational dataset is DIALOCONAN \cite{bonaldi2022human}, which focuses only on hate speech, while similar multi-turn corpora for countering misinformation, either in isolation or intersecting with hate, are not available.

\paragraph{CS generation against hate and misinformation} 
For what regards CS generation against hate, research has shown that language models often produce generic and repetitive
responses, which often fail to address hateful claims effectively \cite{tekiroglu2020generating, munbeyond}. To overcome these limitations, existing studies have focused on improving specific aspects of the generated responses, such as their personalization \citep{de2021toxicbot, dougancc2023generic}, argumentative quality \citep{furman-etal-2023-high, bonaldi2024is}, and factual grounding 
\cite{russo2025trenteam}. In the misinformation domain, CS studies have mainly focused on generating readable, plausible, and faithful responses, evaluated with the involvement of human experts \cite{guo2022survey, russo2023countering, he2023reinforcement}. While earlier approaches relied on attention \citep{popat2018declare} and rule-based methods \cite{gad2019exfakt}, more recent systems  adopt summarization \citep{russo2023benchmarking}, prompting \cite{russo2025euroverdict}, and RAG techniques \cite{russo2025face}.

\section{Data collection methodology}
We collaborated with 23 CS experts (11 fact-checkers and 12 NGO operators) over a period of 18 months\footnote{The entire data collection process included also knowledge search, selection, annotation and translation.} to collect fictitious multi-turn dialogues addressing the co-occurrence of hate and misinformation through a human-machine collaboration setup. In the following sections, we describe the collection process for the external knowledge and the dialogues, and the human-machine collaboration strategies used to create them.

\subsection{External knowledge}
Each dialogue is grounded in one or two external knowledge documents, always including at least one fact-checking article. Below we describe the collection of fact-checking articles and NGO reports, and the matching process for two-document dialogues.
Our data collection is focused on sources that refer to one of the following marginalized groups: Muslims, the LGBTQIA+ community, migrants, women, people with disabilities, people of color and Jewish people. 

\paragraph{Fact-checking articles} \label{subsec:fc_articles_collection}
We use the English fact-checking articles from \citet{martone2026assisted}, and collect new articles by translating their proposed set of keywords
into Polish, Spanish, and Italian using \texttt{deepl}.\footnote{\url{https://pypi.org/project/deepl/}} Then, the translated keywords were reviewed by the experts, who made additions, deletions and substitutions in their respective languages to ensure linguistic coverage of the target minorities and contextual accuracy of the provided terms, resulting in a list of more than 221 keywords spanning the 5 languages (see Table \ref{tab:keywords} in Appendix \ref{appendix:doc_collection} for the full list). Finally, the validated keywords were used to scrape Google Fact Check Tools Explorer\footnote{\url{https://toolbox.google.com/factcheck/explorer/search/list:recent;hl=}} with \texttt{newspaper4k}\footnote{\url{https://pypi.org/project/newspaper4k/}}.
The resulting 2,682 articles underwent a manual filtering aimed at keeping only content (i) written by signatories of the International Fact-Checking Network code of principles\footnote{\url{https://ifcncodeofprinciples.poynter.org/}}, and (ii) that could be used to fuel discrimination against the target groups (see the selection criteria in Appendix \ref{appendix:doc_collection}). Because no dedicated fact-checking sources exist for Maltese, English articles were used as external knowledge to generate Maltese CS, a decision justified by the bilingualism in Malta.
Overall, we collected 516 fact-checking articles across five languages.

\paragraph{NGO reports}
We asked the
NGO operators to help us enrich the set of myth--anti-stereotype pairs created by \citet{martone2026assisted}, by manually searching for additional NGO reports in Polish, Spanish, and Italian (see Appendix \ref{appendix:doc_collection} for the list of domains used). Because the number of non-English reports was limited, we further expanded the dataset by translating and manually checking from the English corpus
specific pairs that were untied to a defined 
country/regional setting.
Also in this case, English documents were used as external knowledge for Maltese dialogues. Overall, we obtained 345 stereotype and anti-stereotype pairs.

\begin{table*}[ht]
\small
\centering
\resizebox{\textwidth}{!}{
\begin{tabular}{ccccccccccccc}
\toprule
\textbf{Docs} & \textbf{MSD$_{or}$} & \textbf{MSD$_{ed}$} & \textbf{$\text{MSD}_{\Delta}$} & \textbf{ASD$_{or}$} & \textbf{ASD$_{ed}$} & \textbf{$\text{ASD}_{\Delta}$} & \textbf{NST$_{or}$} & \textbf{NST$_{ed}$} & \textbf{$\text{NST}_{\Delta}$} & \textbf{CW$_{or}$} & \textbf{CW$_{ed}$} & \textbf{$\text{CW}_{\Delta}$} \\
\midrule
1 & \textbf{4.958} & \textbf{4.923} & -0.035 & \textbf{4.106} & \textbf{4.045} & -0.061 & 2.004 & 2.139 & 0.135 & 0.306 & 0.292 & \textbf{-0.014} \\
2 & 4.601 & 4.763 & \textbf{0.162} & 3.774 & 3.869 & \textbf{0.095} & \textbf{2.071} & \textbf{2.209} & \textbf{0.138} & \textbf{0.346} & \textbf{0.319} & -0.027 \\
\bottomrule
\end{tabular}
}
\caption{Syntactic metrics results according to number of documents used as reference by the dialogues.}
\label{tab:nr_docs_syntactic}
\end{table*}

\paragraph{External knowledge matching} As a final step, we constructed NGO-based external knowledge for the two-document dialogue setting. Unlike fact-checking sources, which consist of coherent articles, NGO materials are structured as isolated myth--anti-stereotype pairs. To align these formats, we computed the semantic similarity\footnote{\texttt{all-mpnet-base-v2} \cite{all-mpnet-base-v2}} between each fact-checking claim and all stereotypes targeting the same group. The resulting matches were manually reviewed, and only meaningful associations were retained. For each fact-checking article, the validated myth--anti-stereotype pairs were then merged into a single document (we refer to it as \textit{NGO report}). This process yielded 313 aligned fact-checking and NGO report pairs.

\subsection{Dialogue collection} \label{subsec:annotation_guidelines}
The collection was performed on the First-AID annotation platform \cite{menini2025first}, designed for annotating knowledge-based dialogues. 
Annotators were tasked to collect multi-turn fictitious dialogues simulating a person spreading hate and misinformation and an operator providing CS grounded on the provided document(s) to counter it. 
Below, we describe (i) the expert annotators, (ii) the guidelines employed to collect the dialogue turns and (iii) the external knowledge annotations.

\paragraph{Expert reviewers} 
The annotation team consisted of 23 CS experts (11 fact-checkers and 12 NGO operators) from Poland, Italy, Malta, and Spain. To protect annotator well-being, we adapted the guidelines of \citet{vidgen2020directions}: the annotation process was distributed over approximately 18 months to avoid excessive workload, and biweekly meetings were held to discuss concerns or difficulties. No major issues emerged, likely due to the annotators’ extensive prior experience with harmful content.

\paragraph{Guidelines: Dialogue turns} 
Annotators were instructed to ensure naturalness and avoid repetition, producing 4–8 turn dialogues starting with a hate-and-misinformation message and ending with a CS response.  
CS writing followed \citet{martone2026assisted} guidelines: (i) avoid abusive language and focus on the message rather than the author, (ii) maintain a respectful and empathetic tone, (iii) counter claims using verified facts and statistics with source grounding, and (iv) provide context while discouraging overgeneralization.

\paragraph{Guidelines: External knowledge} Annotators were provided with external knowledge to support drafting CS: all dialogues always included a fact-checking article, plus an NGO report for the two-document condition. In the latter setting, annotators had to reference both documents at least once across the conversation. They also 
linked the specific document spans supporting each turn. This \textit{ground text} was required to contain all information relevant to the response, and was omitted for turns that didn't require grounding, (e.g. clarification questions "\textit{Where did you get this information?}").

\begin{table}[ht]
\small
\centering
\resizebox{\columnwidth}{!}{%
\begin{tabular}{ccccccc}
\toprule
\textbf{Docs} & \textbf{HTER} & \textbf{Time} & \textbf{Ground} & \textbf{RR$_{or}$} & \textbf{RR$_{ed}$} & \textbf{RR $\Delta$} \\
\midrule
1 & \textbf{0.361} & \textbf{151.5} & 32.0 & \textbf{6.411} & \textbf{4.689} & -1.722 \\
2 & 0.368 & 166.4 & \textbf{45.2} & 8.962 & 6.095 & \textbf{-2.867} \\
\bottomrule
\end{tabular}
}
\caption{Results of annotation effort and RR metrics according to number of documents per dialogue.}
\label{tab:nr_docs_annot_effort}
\end{table}

\subsection{Human-machine collaboration strategies}
We employ the three annotation strategies presented by \citet[][pre-compiled, interactive, and manual]{menini2025first} to reduce the burden on annotators. 
On top of these, we also automatically translate post-edited English dialogues to obtain a multilingual parallel portion of the corpus. More details on the models employed to assist human annotations are in Appendix \ref{appendix:collaboration_strategies}.

\paragraph{Pre-compiled} 
Dialogues are first generated using GPT-4o mini \cite{openai2024gpt4omini}, and the ground text is automatically retrieved by chunking the source documents via SaT \cite{frohmann2024segment} and selecting the highest-similarity chunk with BM25 \cite{robertson1995okapi}. Annotators then review the dialogue and retrieved spans, making necessary adjustments according to the guidelines in \S\ref{subsec:annotation_guidelines}.

\paragraph{Interactive} 
With this strategy, a model dynamically generates multiple turn's alternatives based on the dialogue history. 
Annotators then select the best option, modify it if necessary, or write a new response from scratch. Preliminary experiments showed that zero-shot GPT-4o mini was unsuitable, as its safety guardrails frequently triggered refusals when generating hater turns sequentially, while this issue did not occur when generating full dialogues at once in the pre-compiled strategy. To bypass this, we fine-tuned Llama 3.1 8B \cite{grattafiori2024llama} using data collected with the other strategies. For grounding retrieval, source documents were chunked with LlamaIndex\footnote{\url{https://www.llamaindex.ai/}}, and we experimented with both a zero-shot BGE-M3 retriever \cite{bge-m3} and a fine-tuned BGE-v2-M3 reranker \citep[][]{li2023making, chen2024bge}.

\paragraph{Manual} The dialogue is entirely written by the annotator, who also selects the portion of the document(s) to be used as ground text for CS turns. While this strategy is the most demanding, it serves as a comparison to evaluate the quality of the dialogues obtained with the other approaches.

\paragraph{Translation} 
Part of the post-edited English dialogues is automatically translated using SeamlessM4T Large \cite{barrault2023seamlessm4t}. Annotators then review these translations for correctness and fluency without altering the core content. To reduce their workload, the source documents and grounding text remain untouched in English. Additionally, annotators are encouraged to provide context for elements unfamiliar to their national setting, such as foreign institutions or events (e.g., the UK's ``PIP'' disability benefit).

\section{Annotation and Data description} 
We first outline our evaluation metrics, followed by annotation process statistics calculated at dialogue-level according to the number of referenced documents, dialogue language, and annotation strategy. To prevent bias from unequal subset sizes, all results are reported as macro-averages, calculated first within each language-strategy-number of documents combination and then averaged across the dimensions of interest.
Metrics are divided into two main categories: annotation effort and syntactic metrics.\footnote{${or}$ denotes metrics computed on the original dialogue before post-editing, and ${ed}$ on the post-edited dialogue. Bold shows the best value, underlined the second best: highest for ground and syntactic metrics, lowest for all others.} We report monolingual data first, where both dialogue and article are in the same language, to consistently measure post-editing effort across the manual, interactive, and pre-compiled strategies.
Translation-based data is analyzed separately because translation is applied after human post-editing, introducing an additional transformation that modifies the dialogue surface without changing its content. This affects both metrics that measure annotation effort, 
and all metrics calculated on the ``original'' version of the dialogue, since they depend on the specific post-edited dialogue that was picked to be translated. 
Therefore, translation results are reported independently and are not aggregated with monolingual conditions.

\subsection{Evaluation metrics}

\textbf{Human-targeted Translation Edit Rate (HTER)} measures post-editing effort in terms of ``insertion, deletion, and substitution of single words as well as shifts of word sequences'' \cite{snover2006study}, where higher values equate to greater interventions. The 0.4 threshold is used to identify heavily post-edited dialogues \cite{turchi2013coping}.
\vspace{4pt}

\noindent \textbf{Annotation time} is automatically computed by the First-AID platform in terms of seconds needed to edit the entire dialogue: we normalize it by the number of turns of each dialogue.
\vspace{4pt}

\noindent \textbf{Repetition Rate (RR)} quantifies the lexical diversity of a text in terms of non-singleton n-grams \cite{bertoldi2013cache}. We used a sliding window of 1000 terms over 5 random corpus shuffles.
\vspace{4pt}

\noindent \textbf{Syntactic complexity}: we use the \texttt{spacy} syntactic dependency parser to compute three measures of turn-level syntactic complexity: 
(i) the \textit{Maximum Syntactic Depth} (MSD) and (ii)
\textit{Average Syntactic Depth} (ASD)
of each sentence's dependency tree, and (iii) the \textit{Number of Sentences} (NST).\footnote{\url{https://spacy.io/usage/linguistic-features\#dependency-parse}. These metrics are not computed for Maltese, as no model is available for this language.} 
\vspace{4pt}

\noindent \textbf{Ground length}: for each dialogue, we concatenate all the annotated ground text and compute the average words-level length.

\subsection{Monolingual data}
\paragraph{Number of documents}

Table \ref{tab:nr_docs_annot_effort} shows that single-document dialogues require less annotation effort, yielding lower HTER and shorter average annotation times. While their RR is also lower, post-editing successfully reduces repetitiveness across both configurations. As expected, single-document dialogues have shorter average grounding lengths due to more limited reference knowledge. Regarding syntactic metrics (Table \ref{tab:nr_docs_syntactic}), single-document dialogues exhibit higher MSD and ASD, despite containing fewer sentences on average, alongside a lower proportion of complex words.

\begin{table}
\small
\centering
\resizebox{\columnwidth}{!}{%
\begin{tabular}{lrrrrrr}
\toprule
\textbf{Lang.} & \textbf{HTER} & \textbf{Time} & \textbf{Ground} & \textbf{RR$_{or}$} & \textbf{RR$_{ed}$} & \textbf{RR $\Delta$} \\
\midrule
EN & 0.508 & 286.118 & 21.280 & {\ul 6.156} & {\ul 3.418} & {\ul -2.738} \\
ES & \textbf{0.110} & {\ul 96.352} & \textbf{50.764} & 9.622 & 8.269 & -1.353 \\
IT & {\ul 0.205} & \textbf{67.578} & 38.938 & 9.243 & 6.980 & -2.263 \\
MT & 0.507 & 139.948 & 25.500 & \textbf{5.215} & 3.699 & -1.516 \\
PL & 0.562 & 221.050 & {\ul 45.106} & 6.792 & \textbf{3.355} & \textbf{-3.437} \\
\bottomrule
\end{tabular}
}
\caption{Results of annotation effort and RR metrics according to dialogue language.}
\label{tab:language_annot_effort}
\end{table}

\begin{table*}
\small
\centering
\resizebox{\textwidth}{!}{%
\begin{tabular}{lcccccccccccc}
\toprule
\textbf{Lang.} & \textbf{MSD$_{or}$} & \textbf{MSD$_{ed}$} & \textbf{$\text{MSD}_{\Delta}$} & \textbf{ASD$_{or}$} & \textbf{ASD$_{ed}$} & \textbf{$\text{ASD}_{\Delta}$} & \textbf{NST$_{or}$} & \textbf{NST$_{ed}$} & \textbf{$\text{NST}_{\Delta}$} & \textbf{CW$_{or}$} & \textbf{CW$_{ed}$} & \textbf{$\text{CW}_{\Delta}$} \\
\midrule
EN & \textbf{6.156} & \textbf{5.548} & -0.608 & \textbf{5.010} & \textbf{4.443} & -0.567 & \textbf{2.248} & {\ul 2.218} & -0.030 & 0.178 & 0.140 & -0.038 \\
ES & 4.531 & 4.592 & 0.061 & 3.769 & 3.864 & \textbf{0.095} & 1.982 & 1.966 & -0.016 & 0.311 & 0.303 & -0.008 \\
IT & 4.532 & 4.629 & {\ul 0.097} & {\ul 3.816} & {\ul 3.866} & 0.050 & 1.969 & 2.063 & {\ul 0.094} & {\ul 0.362} & {\ul 0.357} & {\ul -0.005} \\
MT & - & - & - & - & - & - & - & - & - & 0.292 & 0.270 & -0.022 \\
PL & {\ul 4.676} & {\ul 4.822} & \textbf{0.146} & 3.781 & 3.839 & {\ul 0.058} & {\ul 2.038} & \textbf{2.423} & \textbf{0.385} & \textbf{0.403} & \textbf{0.406} & \textbf{0.003} \\
\bottomrule
\end{tabular}
}
\caption{Syntactic metrics results according to dialogue language.}
\label{tab:language_syntactic}
\end{table*}

\paragraph{Language}
Tables \ref{tab:language_annot_effort} and \ref{tab:language_syntactic} present the annotation effort and syntactic metrics by language. Spanish and Italian require the least effort, yielding the lowest average annotation times and the only HTER below 0.4. The higher HTER in other languages stems from distinct factors. On the one hand Polish and Maltese are low-resourced languages, which could explain worse quality generations used as a starting point. Meanwhile, English and Polish show significantly higher annotation times (>200 seconds per turn) paired with the largest reductions in RR, suggesting significant effort to fix repetitive generations. Syntactic metrics offer further insight for Polish: it consistently registers the highest or second-highest increase in syntactic complexity with post-editing, showing that Polish annotators frequently split and articulated sentences with respect to the original generation.

\begin{table}
\small
\centering
\resizebox{\columnwidth}{!}{%
\begin{tabular}{lrrrrrr}
\toprule
\textbf{Strat.} & \textbf{HTER} & \textbf{Time} & \textbf{Ground} & \textbf{RR$_{or}$} & \textbf{RR$_{ed}$} & \textbf{RR $\Delta$} \\
\midrule
Interactive & {\ul 0.419} & {\ul 165.286} & {\ul 31.433} & {\ul 10.296} & 7.917 & \textbf{-2.379} \\
Manual & - & 235.52 & 28.068 & - & \textbf{2.536} & - \\
Pre-compiled & \textbf{0.309} & \textbf{96.491} & \textbf{48.372} & \textbf{4.793} & {\ul 4.302} & {\ul -0.491} \\
\bottomrule
\end{tabular}
}
\caption{Results for annotation effort metrics according to the annotation strategy employed.}
\label{tab:strategy_annot_effort}
\end{table}

\begin{table*}
\small
\centering
\resizebox{\textwidth}{!}{%
\begin{tabular}{lrrrrrrrrrrrr}
\toprule
\textbf{Strat.} & \textbf{MSD$_{or}$} & \textbf{MSD$_{ed}$} & \textbf{MSD$_{\Delta}$} & \textbf{ASD$_{or}$} & \textbf{ASD$_{ed}$} & \textbf{ASD$_{\Delta}$} & \textbf{NST$_{or}$} & \textbf{NST$_{ed}$} & \textbf{NST$_{\Delta}$} & \textbf{CW$_{or}$} & \textbf{CW$_{ed}$} & \textbf{CW$_{\Delta}$} \\
\midrule
Interactive & \textbf{5.001} & {\ul 4.793} & {\ul -0.208} & \textbf{4.165} & {\ul 4.015} & {\ul -0.150} & {\ul 1.986} & 2.007 & {\ul 0.021} & \textbf{0.339} & \textbf{0.329} & {\ul -0.010} \\
Manual & - & \textbf{5.148} & - & - & \textbf{4.174} & - & - & {\ul 2.237} & - & - & 0.251 & - \\
Pre-compiled & {\ul 4.609} & 4.731 & \textbf{0.122} & {\ul 3.762} & 3.806 & \textbf{0.044} & \textbf{2.079} & \textbf{2.271} & \textbf{0.192} & {\ul 0.309} & {\ul 0.308} & \textbf{-0.001} \\
\bottomrule
\end{tabular}
}
\caption{Results for syntactic metrics according to the annotation strategy employed.}
\label{tab:strategy_syntactic}
\end{table*}

\paragraph{Annotation strategy}
Tables \ref{tab:strategy_annot_effort} and \ref{tab:strategy_syntactic} break down annotation effort and syntactic metrics by strategy. As shown in Table \ref{tab:strategy_annot_effort}, the pre-compiled strategy requires the least effort, yielding the lowest annotation times and the only HTER below 0.4, while producing the longest ground text. Conversely, the manual strategy demands the most time but results in the lowest RR$_{ed}$. This suggests that while synthetic baselines drastically reduce annotation time, they partially bias annotators toward specific word choices. Consequently, machine-assisted strategies increase dialogue repetitiveness and fail to match the lexical diversity of text written by humans entirely from scratch. The interactive strategy exhibits the highest RR${ed}$, yet annotators invest the most effort here into reducing repetition, as shown by the high RR $\Delta$. The RR$_{or}$ indicates that the model generates more varied content when tasked with producing the entire dialogue at once rather than interactively. This occurs in the pre-compiled strategy, where the model can plan its generation in advance, rather than generating turn-by-turn, where it exhibits a greedier behavior and falls more easily into repetitive phrasing. Regarding syntactic metrics (Table \ref{tab:strategy_syntactic}), manual dialogues have the highest MSD and ASD alongside the lowest proportion of complex words, indicating a complex sentence structure paired with common vocabulary. In contrast, the pre-compiled strategy, across both generated and post-edited versions, features the lowest MSD and ASD but the highest NST, reflecting a simpler syntax where content is split across more sentences. Results grouped by number of documents and strategy are coherent with those shown by strategy (Tables \ref{tab:docs_strategy_annot_effort} and \ref{tab:docs_strategy_syntactic} in Appendix \ref{app:more_monolingual_analyses}).

\begin{table}
\small
\centering
\resizebox{\columnwidth}{!}{%
\begin{tabular}{lrrrrr}
\toprule
\textbf{Lang.} & \textbf{HTER} & \textbf{Time} & \textbf{RR$_{or}$} & \textbf{RR$_{ed}$} & \textbf{RR $\Delta$} \\
\midrule
ES & \textbf{1.014} & {\ul 85.350} & 3.141 & 4.210 & 1.069 \\
IT & {\ul 1.039} & \textbf{30.795}& 2.704 & 3.460 & 0.756 \\
MT & 1.422 & 179.995 & \textbf{2.156} & {\ul 2.148} & {\ul -0.008} \\
PL & 1.322 & 133.775  & {\ul 2.172} & \textbf{1.974} & \textbf{-0.198} \\
\bottomrule
\end{tabular}
}
\caption{Annotation effort results for translated dialogues according to the language.}
\label{tab:trans_lang_annot_effort}
\end{table}

\begin{table*}
\small
\centering
\resizebox{\textwidth}{!}{%
\begin{tabular}{lrrrrrrrrrrrr}
\toprule
\textbf{Lang.} & \textbf{MSD$_{or}$} & \textbf{MSD$_{ed}$} & \textbf{$\text{MSD}_{\Delta}$} & \textbf{ASD$_{or}$} & \textbf{ASD$_{ed}$} & \textbf{$\text{ASD}_{\Delta}$} & \textbf{NST$_{or}$} & \textbf{NST$_{ed}$} & \textbf{$\text{NST}_{\Delta}$} & \textbf{CW$_{or}$} & \textbf{CW$_{ed}$} & \textbf{$\text{CW}_{\Delta}$} \\
\midrule
ES & {\ul 4.766} & 4.82 & {\ul 0.054} & {\ul 3.974} & \textbf{3.948} & -0.026 & {\ul 2.044} & 2.112 & {\ul 0.068} & 0.309 & 0.308 & {\ul -0.001} \\
IT & \textbf{4.847} & {\ul 4.828} & -0.019 & 3.913 & 3.836 & {\ul -0.077} & \textbf{2.172} & \textbf{2.224} & 0.052 & {\ul 0.361} & {\ul 0.360} & {\ul -0.001} \\
MT & - & - & - & - & - & - & - & - & - & 0.252 & 0.254 & 0.002 \\
PL & 4.75 & \textbf{4.866} & \textbf{0.116} & \textbf{4.084} & {\ul 3.928} & \textbf{-0.156} & 1.874 & {\ul 2.218} & \textbf{0.344} & \textbf{0.422} & \textbf{0.417} & \textbf{-0.005} \\
\bottomrule
\end{tabular}
}
\caption{Syntactic metrics results for translated dialogues according to the language}
\label{tab:trans_lang_syntactic}
\end{table*}

\subsection{Translated data} \label{subsec:translated_data}
Tables \ref{tab:trans_lang_annot_effort} and \ref{tab:trans_lang_syntactic} report the metrics for translated dialogues.\footnote{Ground length is excluded as it was not modified during annotation.} Across all languages, HTER exceeds 1.0, indicating low-quality automatic translations that required substantial post-editing. Spanish and Italian demanded the least effort (lowest HTER and annotation time), yet their RR increased post-annotation. Their syntactic metrics remained largely stable (deltas near 0.0), suggesting the process did not alter the dialogues' core structure. This indicates that because the initial Spanish and Italian translations were relatively fluent, annotators made shallow, local edits rather than deep restructurings. This minimal patching preserved or amplified the repetitive phrasing already present in the machine translation. Conversely, for lower-resourced languages like Polish and Maltese, annotators were forced to reformulate text aggressively. This deeper intervention resulted in higher HTER, longer annotation times, larger syntactic deltas (MSD, ASD, and NST for Polish), and a corresponding reduction in repetition. These patterns are consistent when grouping results by the number of documents and language (see Tables \ref{tab:trans_docs_lang_annot_effort} and \ref{tab:trans_docs_lang_syntactic} in Appendix \ref{appendix:translation_analyses}).

\subsection{Final dataset description}
The final dataset contains 2,015 dialogues (1,565 single-document, 450 two-document) and 12,298 turns. 
The distribution across languages is uniform, with each accounting for 14--16\% of single-document and 2.5--5\% of two-document dialogues. Conversely, the distribution of targeted groups is skewed due to the varying real-world availability of fact-checking articles addressing specific marginalised groups. Dialogues predominantly focus on Migrants (30\%), Muslims (19\%), and Women (17\%); despite targeted scraping efforts, dialogues concerning 
Jewish people and individuals with disabilities are a minority (2--7\%): rather than artificially balancing the corpus with suboptimal examples, we prioritize cases for which experts could rely on verifiable evidence.  The distribution of dialogues according to target and language is available in Tables \ref{tab:target_coverage} \ref{tab:lang_coverage}, and \ref{tab:target_language_coverage} in Appendix \ref{appendix:data_description}.

\section{Experiments}
In the following experiments, we test how our dataset
can be used to evaluate retrieval and generation in fact-based multilingual CS generation settings when faced with overlapping hate and misinformation. To achieve this, we conduct two main experiments: a retrieval and a generation task.

\begin{table}[ht]
\small
\centering
\resizebox{\columnwidth}{!}{%
\begin{tabular}{llrrr}
\toprule
\textbf{Model} & \textbf{Query} & \textbf{Hit@10} & \textbf{MAP@10} & \textbf{Recall@10} \\
\midrule
\multirow{3}{*}{BM25} 
 & Q        & 0.279 & 0.135 & 0.235 \\
 & Q$_{\text{DC}}$ & \textbf{0.452} & \textbf{0.218} & \textbf{0.397} \\
 & Q$_{\text{R}}$  & 0.425 & 0.209 & 0.369 \\
\midrule
\multirow{3}{*}{BGE-M3} 
 & Q        & 0.384 & 0.192 & 0.341 \\
 & Q$_{\text{DC}}$ & \textbf{0.550} & 0.272 & \textbf{0.500} \\
 & Q$_{\text{R}}$  & 0.540 & \textbf{0.273} & 0.491 \\
\midrule
\multirow{3}{*}{Qwen3} 
 & Q        & 0.464 & 0.238 & 0.418 \\
 & Q$_{\text{DC}}$ & \textbf{*0.595} & \textbf{*0.306} & \textbf{*0.548} \\
 & Q$_{\text{R}}$  & 0.579 & 0.305 & 0.531 \\
\bottomrule
\end{tabular}}
\caption{Monolingual zero-shot chunk retrieval performance across varying query configurations.}
\label{tab:retr_monolingual}
\end{table}

\subsection{Retrieval} \label{subsec:retrieval}
We evaluate three embedders on a zero-shot chunk retrieval task for hate and misinformation countering, across three query configurations and two sub-tasks: monolingual and cross-lingual retrieval.\footnote{More details on the preprocessing, prompts, and hyperparameters are provided in Appendix \ref{appendix:retrieval_details}.}

\paragraph{Retriever} 
We compare BM25 as a sparse baseline against two dense retrievers representing distinct structural paradigms: BGE-M3, a dedicated encoder-only model engineered for multi-functional retrieval, and Qwen3-Embedding-4B \cite{qwen3embedding}, a modern LLM-based embedder. Despite both models offer multilingual support, Qwen3-Embedding is the only offering native support for the Maltese language.

\paragraph{Query} 
We test three different query configurations: (i) \textbf{Original query} ($Q$): each hate speech turn is used as is to retrieve the relevant document chunks needed to formulate the CS; (ii) \textbf{Dialogue context query} ($Q_{DC}$): the query is formed by concatenating the original query with all preceding conversation turns \cite{wu2022conqrr}; (iii) \textbf{Rewritten query} ($Q_R$): for all hate speech turns that are not the first, the query is rewritten by conditioning it on the preceding dialogue context using the prompt template from \citet{ye2023enhancing}.
    
\paragraph{Task} 
Retrieval performance is measured in two settings: (i) \textbf{monolingual}: queries and reference documents share the same language; (ii) \textbf{cross-lingual}: the reference document is in English, while queries are in a different language.
In both configurations, the search space contains all chunks from all source documents.

\paragraph{Results}
Table \ref{tab:retr_monolingual} presents the zero-shot monolingual retrieval results\footnote{Bold text indicates the best performance within each model group. * denotes the overall best for each metric.}: cross-lingual results are consistent and reported in Table \ref{tab:retr_crosslingual} (Appendix \ref{appendix:retrieval_details}). Two distinct trends emerge. First, dense models substantially outperform the lexical baseline: Qwen3-Embedding consistently achieves the highest performance, followed by BGE-M3 and BM25, across all settings. Second, query formulation heavily impacts success. Across all models, providing the full dialogue context ($Q_{\text{DC}}$) yields the best performance, followed closely by the LLM-rewritten query ($Q_{\text{R}}$), while the standalone query ($Q$) performs worst. This underscores the necessity of conversational context to resolve implicit or vague references typical in dialogue-based hate speech and misinformation. A key divergence occurs in the cross-lingual scenario: while Qwen3 maintains its preference for $Q_{\text{DC}}$, both BM25 and BGE-M3 perform better with $Q_{\text{R}}$. This variation suggests that explicit query rewriting may filter out conversational noise for BM25 and BGE-M3 during cross-lingual transfer, whereas larger models like Qwen3 can align dense, multilingual representations directly from raw conversational context.

\begin{table}[ht]
\small
\centering
\begin{tabular}{llr}
\toprule
\textbf{Metric} & \textbf{Setting} & \textbf{Value} \\
\midrule
\multirow{3}{*}{BERTScore} 
 & CS$_{base}$         & 0.884 \\
 & CS$_{gold}$ & \textbf{0.895} \\
 & CS$_{retr}$ & 0.888 \\
 
\midrule
\multirow{3}{*}{Faithfulness$_{gold}$} 
 & CS$_{base}$              & 2.702 \\
 & CS$_{gold}$ & \textbf{4.116} \\
 & CS$_{retr}$ & 3.138 \\
 \midrule
Faithfulness$_{retr}$ & CS$_{retr}$ & \textbf{4.496} \\
\midrule
\multirow{3}{*}{NLI Entailment$_{gold}$} 
 & CS$_{base}$         & 0.033 \\
 & CS$_{gold}$ & \textbf{0.246} \\
 &          CS$_{retr}$ &  0.086 \\
 \midrule
 NLI Entailment$_{retr}$ & CS$_{retr}$ & 0.228 \\
 
\midrule
\multirow{3}{*}{Relevance} 
 & CS$_{base}$         & \textbf{4.711} \\
 & CS$_{gold}$ & 4.538 \\
 & CS$_{retr}$ & 4.514 \\
 
\bottomrule
\end{tabular}
\caption{CS generation quality across metrics.}
\label{tab:gen_results_mono}
\end{table}

\subsection{Generation}
We employ Qwen3 8B \cite{qwen3technicalreport} to test three zero-shot configurations for CS generation, with different input: (i) CS$_{base}$: only the harmful statement containing hate and misinformation (and dialogue history, if present), (ii) CS$_{gold}$: CS$_{base}$ plus the gold knowledge, and (iii) CS$_{retr}$: CS$_{base}$ plus the top 5 chunks retrieved by Qwen3-Embedding, i.e. the best retriever emerged from \S\ref{subsec:retrieval} (See Appendix \ref{appendix:generation_details} for more details). 

\paragraph{Evaluation metrics}
We evaluate the generated CS using four metrics: \textbf{BERTScore} to capture semantic similarity with the human-curated CS \cite{zhang2019bertscore}\footnote{We report $F_1$ scores using \texttt{xlm-roberta-large} \cite{conneau2020unsupervised}};
\textbf{NLI Entailment} to assesse factual alignment by computing the probability that the output is entailed by the source text via \texttt{xlm-roberta-large-xnli} \cite{joeddav2020xlmrobertaxnli}; 
\textbf{Faithfulness} to evaluate adherence to the context without hallucinations;
\textbf{Relevance} to measure how appropriately the CS addresses the previous turn. We use GPT-4.1 mini \cite{openai_gpt41mini_2025} to score Faithfulness and Relevance on a 1–5 Likert scale, adapting the definitions from \citet{es2024ragas}. For $\text{CS}_{\text{retr}}$, NLI and Faithfulness are computed against both gold and retrieved knowledge.

\paragraph{Results}
As shown in Table \ref{tab:gen_results_mono}, $\text{CS}_{\text{gold}}$ serves as the optimal generation configuration across all metrics except relevance, followed closely by $\text{CS}_{\text{retr}}$, while the ungrounded baseline (CS$_{base}$) consistently performs worst. CS$_{base}$ yields the lowest semantic similarity to human-curated CS. Providing retrieved chunks elevates the BERTScore to $0.888$, closely matching $\text{CS}_{\text{gold}}$ ($0.895$) and demonstrating that automated retrieval successfully improves the generation's alignment to expert-curated CS. For NLI Entailment and Faithfulness, CS$_{base}$ exhibits a substantial misalignment with the verified facts, yielding the lowest scores ($0.033$ and $2.702$). On the other hand, as regards CS$_{retr}$, when it is evaluated against the retrieved context, its performance approaches or even surpasses CS$_{gold}$ on NLI Entailment and Faithfulness, respectively. However, evaluating those same generations against the hidden gold knowledge causes a performance drop. While this remains a major improvement over CS$_{base}$, it underscores that generator quality relies heavily on retriever accuracy. Finally, CS$_{base}$ achieves the highest relevance score ($4.711$), outperforming $\text{CS}_{\text{retr}}$ ($4.514$) and $\text{CS}_{\text{gold}}$ ($4.538$). This result highlights an inherent trade-off in evidence-grounded generation: forcing a model to integrate external evidence slightly restrains its ability to adhere to the user query. However, this is an acceptable trade-off when the objective is to counter hate speech and misinformation with verifiable facts. The same dynamics and trends are observed in the cross-lingual configuration (see Table \ref{tab:gen_results_cross} in Appendix \ref{appendix:generation_details}).

\section{Conclusion}
While prior NLP resources treat hate and misinformation
in isolation, we introduce CATCH-ME: the first multilingual, multi-turn counterspeech dataset designed to tackle intertwined toxic narratives and false claims simultaneously. Developed with 23 domain experts across four human-machine collaboration strategies, our corpus provides high-quality refutations in five languages, fully grounded in fact-checking articles and NGO reports. Beyond introducing this resource, we analyze the human-in-the-loop annotation process and establish robust strong retrieval and generation benchmarks baselines in RAG settings. Ultimately, this dataset provides a rigorous framework for training fact-based counterspeech models to foster safer online discourse.

\section*{Acknowledgements}
This work was partially supported by the European Union’s CERV fund under grant agreement No. 101143249 (HATEDEMICS). We are grateful to the following NGOs, fact-checking organizations and all annotators for their help: ALDA (Association Europeenne Pour La Democratie Locale), FUNDEA (Fundacion Euroarabe De Altos Estudios), MALDITA (Fundacion Maldita.Es Contra la Desinformacion: Periodismo educacion Investigacion y Datos En Nuevos Formatos), CENTRA (Fundacion Pública Andaluza Centro De Estudios Andaluces M.P.), CESIE ETS (CESIE ETS), TFCF (The Fact-Checking Factory S.r.l.), SOS MALTA (Solidarity And Overseas Service Malta), VSA (Victim Support Agency), CEO (Fundacja Centrum Edukacji Obywatelskiej), DEMAGOG (Stowarzyszenie Demagog), NASK (Naukowa I Akademicka Siec Komputerowa - Panstwowy Instytut Badawczy).

\section*{Limitations}
Our dataset is multilingual by design, but not intended to exhaustively cover the linguistic diversity of online hate and misinformation. We include both higher-resource languages, i.e., English, Italian, and Spanish, and lower-resource languages, i.e., Polish and Maltese, as a feasible compromise between breadth, expert availability, and the need for high-quality grounding and post-editing. Future work can build on this resource by extending the same methodology to additional languages, especially underrepresented linguistic communities.

Similarly, the dataset covers multiple targets of hate, but their distribution is not uniform. This reflects the availability of reliable external knowledge: some groups are more frequently represented in fact-checking articles and anti-stereotype resources than others. Rather than artificially balancing the corpus with suboptimal examples, we prioritize cases for which experts could rely on verifiable evidence. Future expansions can address this imbalance through targeted collection efforts and collaborations with organizations specializing in less represented communities.

The hateful and misinformed turns in our dialogues are fictitious. This choice allows annotators to focus on the main objective of the resource: writing grounded counterspeech that jointly addresses false claims and harmful stereotypes. Consequently, the dataset should not be interpreted as a comprehensive model of how hate speech and misinformation appear in naturally occurring online interactions, where they may be more implicit, ambiguous, or context-dependent. Future research can use our corpus as a controlled starting point and evaluate transfer to naturally occurring conversations.

Finally, our retrieval and generation experiments are intended as initial benchmarks rather than an exhaustive evaluation of all possible modeling choices. Since the main contribution of this work is the collection of a large expert-curated, knowledge-grounded dialogue dataset, we test a limited set of retrieval and generation models and rely on automatic evaluation metrics. These experiments establish a first reference point for the task enabled by our dataset. Future work can extend this benchmark with additional model families, fine-tuning strategies, and human evaluation of factuality, persuasiveness, safety, and conversational appropriateness.

\section*{Ethical consideration}
This work addresses harmful online content and therefore requires explicit safeguards for both the people involved in data creation and the potential downstream use of the resource. 

\paragraph{Annotator well-being.}
The dataset was created with domain experts who had prior experience with counterspeech, fact-checking, and anti-discrimination work. To reduce the burden of repeated exposure to offensive and misleading content, the annotation campaign was distributed over an extended period and included regular meetings where annotators could discuss difficulties, raise concerns, and receive support. The annotation guidelines also required counterspeech to avoid abusive language, address the content rather than the author, and maintain a respectful and empathetic tone. Finally, annotators were compensated fairly by their respective institutions in compliance with applicable national laws.

\paragraph{Privacy and data provenance.}
The dialogues in the dataset are fictitious and were produced through expert writing, post-editing, and human-machine collaboration rather than by scraping conversations between real users. This choice avoids the collection of personal user interactions and reduces privacy risks associated with sensitive online discussions. At the same time, the counterspeech turns are grounded in external knowledge from fact-checking articles and NGO-derived anti-stereotype material, so that responses are based on verifiable evidence rather than personal information. To comply with licensing requirements, we will not redistribute the original external knowledge text, but we will make available the code to replicate our data collection and to obtain the span-level annotations from the web pages links.

\paragraph{Harmful content and misuse.}
Because the dataset targets the intersection of hate speech and misinformation, it necessarily contains offensive and misleading statements. We keep such statements explicit and stereotypical to support the controlled study of grounded counterspeech generation, not to model or amplify realistic hateful behavior. Moreover, dialogues are structured so that the final turn is always a counterspeech response, reducing the risk of presenting hateful or misleading content as the conversational endpoint. The dataset is intended for research on safer counterspeech generation, retrieval, and evaluation, and should not be used to train systems that generate, rank, or amplify hateful or misleading content.

\paragraph{Model use.}
The generation experiments in this paper are intended as benchmarks for the task enabled by the dataset, not as deployable moderation or intervention systems. Automatically generated counterspeech can be factually incomplete, contextually inappropriate, or ineffective in sensitive real-world settings. For this reason, we view the models evaluated here as tools for research and for assisting expert data creation, rather than as replacements for trained moderators, fact-checkers, or civil-society practitioners. Any deployment-oriented use should include human oversight, additional safety evaluation, and context-specific validation.

\bibliography{anthology,custom}

\begin{thebibliography}{59}
\providecommand{\natexlab}[1]{#1}

\bibitem[{Barrault et~al.(2023)Barrault, Chung, Meglioli, Dale, Dong, Duquenne, Elsahar, Gong, Heffernan, Hoffman et~al.}]{barrault2023seamlessm4t}
Lo{\"\i}c Barrault, Yu-An Chung, Mariano~Cora Meglioli, David Dale, Ning Dong, Paul-Ambroise Duquenne, Hady Elsahar, Hongyu Gong, Kevin Heffernan, John Hoffman, and 1 others. 2023.
\newblock Seamlessm4t: Massively multilingual \& multimodal machine translation.
\newblock \emph{arXiv preprint arXiv:2308.11596}.

\bibitem[{Bertoldi et~al.(2013)Bertoldi, Cettolo, and Federico}]{bertoldi2013cache}
Nicola Bertoldi, Mauro Cettolo, and Marcello Federico. 2013.
\newblock Cache-based online adaptation for machine translation enhanced computer assisted translation.
\newblock In \emph{MT-Summit}, pages 35--42.

\bibitem[{Bonaldi et~al.(2024{\natexlab{a}})Bonaldi, Chung, Abercrombie, and Guerini}]{bonaldi-etal-2024-nlp}
Helena Bonaldi, Yi-Ling Chung, Gavin Abercrombie, and Marco Guerini. 2024{\natexlab{a}}.
\newblock \href {https://doi.org/10.18653/v1/2024.findings-naacl.221} {{NLP} for counterspeech against hate: A survey and how-to guide}.
\newblock In \emph{Findings of the Association for Computational Linguistics: NAACL 2024}, pages 3480--3499, Mexico City, Mexico. Association for Computational Linguistics.

\bibitem[{Bonaldi et~al.(2024{\natexlab{b}})Bonaldi, Damo, Ocampo, Cabrio, Villata, and Guerini}]{bonaldi2024is}
Helena Bonaldi, Greta Damo, Nicol{\'a}s~Benjam{\'i}n Ocampo, Elena Cabrio, Serena Villata, and Marco Guerini. 2024{\natexlab{b}}.
\newblock \href {https://doi.org/10.18653/v1/2024.emnlp-main.201} {Is safer better? the impact of guardrails on the argumentative strength of {LLM}s in hate speech countering}.
\newblock \emph{Proceedings of the 2024 Conference on Empirical Methods in Natural Language Processing}, pages 3446--3463.

\bibitem[{Bonaldi et~al.(2022)Bonaldi, Dellantonio, Tekiro{\u{g}}lu, and Guerini}]{bonaldi2022human}
Helena Bonaldi, Sara Dellantonio, Serra~Sinem Tekiro{\u{g}}lu, and Marco Guerini. 2022.
\newblock Human-machine collaboration approaches to build a dialogue dataset for hate speech countering.
\newblock In \emph{Proceedings of the 2022 Conference on Empirical Methods in Natural Language Processing}, pages 8031--8049.

\bibitem[{Cazzamatta(2025)}]{cazzamatta2025global}
Regina Cazzamatta. 2025.
\newblock Global misinformation trends: Commonalities and differences in topics, sources of falsehoods, and deception strategies across eight countries.
\newblock \emph{new media \& society}, 27(11):6334--6358.

\bibitem[{Chen et~al.(2024{\natexlab{a}})Chen, Xiao, Zhang, Luo, Lian, and Liu}]{bge-m3}
Jianlv Chen, Shitao Xiao, Peitian Zhang, Kun Luo, Defu Lian, and Zheng Liu. 2024{\natexlab{a}}.
\newblock \href {https://arxiv.org/abs/2402.03216} {Bge m3-embedding: Multi-lingual, multi-functionality, multi-granularity text embeddings through self-knowledge distillation}.
\newblock \emph{Preprint}, arXiv:2402.03216.

\bibitem[{Chen et~al.(2024{\natexlab{b}})Chen, Xiao, Zhang, Luo, Lian, and Liu}]{chen2024bge}
Jianlv Chen, Shitao Xiao, Peitian Zhang, Kun Luo, Defu Lian, and Zheng Liu. 2024{\natexlab{b}}.
\newblock \href {https://arxiv.org/abs/2402.03216} {Bge m3-embedding: Multi-lingual, multi-functionality, multi-granularity text embeddings through self-knowledge distillation}.
\newblock \emph{Preprint}, arXiv:2402.03216.

\bibitem[{Chung et~al.(2019)Chung, Kuzmenko, Tekiro{\u{g}}lu, and Guerini}]{conan-2019}
Yi-Ling Chung, Elizaveta Kuzmenko, Serra~Sinem Tekiro{\u{g}}lu, and Marco Guerini. 2019.
\newblock \href {https://doi.org/10.18653/v1/P19-1271} {{CONAN} - {CO}unter {NA}rratives through nichesourcing: a multilingual dataset of responses to fight online hate speech}.
\newblock In \emph{Proceedings of the 57th Annual Meeting of the Association for Computational Linguistics}, pages 2819--2829, Florence, Italy. Association for Computational Linguistics.

\bibitem[{Chung et~al.(2021{\natexlab{a}})Chung, Tekiro{\u{g}}lu, and Guerini}]{chung2021towards}
Yi-Ling Chung, Serra~Sinem Tekiro{\u{g}}lu, and Marco Guerini. 2021{\natexlab{a}}.
\newblock \href {https://doi.org/10.18653/v1/2021.findings-acl.79} {Towards knowledge-grounded counter narrative generation for hate speech}.
\newblock \emph{Findings of the Association for Computational Linguistics: ACL-IJCNLP 2021}, pages 899--914.

\bibitem[{Chung et~al.(2021{\natexlab{b}})Chung, Tekiro{\u{g}}lu, Tonelli, and Guerini}]{chung2021empowering}
Yi-Ling Chung, Serra~Sinem Tekiro{\u{g}}lu, Sara Tonelli, and Marco Guerini. 2021{\natexlab{b}}.
\newblock Empowering {NGO}s in countering online hate messages.
\newblock \emph{Online Social Networks and Media}, 24:100150.

\bibitem[{Conneau et~al.(2020)Conneau, Khandelwal, Goyal, Chaudhary, Wenzek, Guzm{\'a}n, Grave, Ott, Zettlemoyer, and Stoyanov}]{conneau2020unsupervised}
Alexis Conneau, Kartikay Khandelwal, Naman Goyal, Vishrav Chaudhary, Guillaume Wenzek, Francisco Guzm{\'a}n, Edouard Grave, Myle Ott, Luke Zettlemoyer, and Veselin Stoyanov. 2020.
\newblock Unsupervised cross-lingual representation learning at scale.
\newblock In \emph{Proceedings of the 58th annual meeting of the association for computational linguistics}, pages 8440--8451.

\bibitem[{Davison(2020)}]{joeddav2020xlmrobertaxnli}
Joe Davison. 2020.
\newblock joeddav/xlm-roberta-large-xnli.
\newblock \url{https://huggingface.co/joeddav/xlm-roberta-large-xnli}.

\bibitem[{de~los Riscos and D’Haro(2021)}]{de2021toxicbot}
Agust{\'\i}n~Manuel de~los Riscos and Luis~Fernando D’Haro. 2021.
\newblock Toxicbot: A conversational agent to fight online hate speech.
\newblock \emph{Conversational dialogue systems for the next decade}, pages 15--30.

\bibitem[{Dettmers et~al.(2023)Dettmers, Pagnoni, Holtzman, and Zettlemoyer}]{dettmers2023qlora}
Tim Dettmers, Artidoro Pagnoni, Ari Holtzman, and Luke Zettlemoyer. 2023.
\newblock Qlora: Efficient finetuning of quantized llms.
\newblock \emph{Advances in neural information processing systems}, 36:10088--10115.

\bibitem[{Do{\u{g}}an{\c{c}} and Markov(2023)}]{dougancc2023generic}
Mekselina Do{\u{g}}an{\c{c}} and Ilia Markov. 2023.
\newblock From generic to personalized: Investigating strategies for generating targeted counter narratives against hate speech.
\newblock In \emph{Proceedings of the 1st Workshop on CounterSpeech for Online Abuse (CS4OA)}, pages 1--12.

\bibitem[{Es et~al.(2024)Es, James, Anke, and Schockaert}]{es2024ragas}
Shahul Es, Jithin James, Luis~Espinosa Anke, and Steven Schockaert. 2024.
\newblock Ragas: Automated evaluation of retrieval augmented generation.
\newblock In \emph{Proceedings of the 18th conference of the european chapter of the association for computational linguistics: system demonstrations}, pages 150--158.

\bibitem[{Fanton et~al.(2021)Fanton, Bonaldi, Tekiro{\u{g}}lu, and Guerini}]{fanton2021human}
Margherita Fanton, Helena Bonaldi, Serra~Sinem Tekiro{\u{g}}lu, and Marco Guerini. 2021.
\newblock Human-in-the-loop for data collection: a multi-target counter narrative dataset to fight online hate speech.
\newblock In \emph{Proceedings of the 59th Annual Meeting of the Association for Computational Linguistics and the 11th International Joint Conference on Natural Language Processing (Volume 1: Long Papers)}, pages 3226--3240.

\bibitem[{Fontana(2022)}]{facta2022LBGTmyths}
Simone Fontana. 2022.
\newblock \href {https://www.facta.news/articoli/dieci-falsi-miti-da-sfatare-sulla-comunita-lgbt} {Dieci falsi miti da sfatare sulla comunità lgbt+}.
\newblock Accessed: 2026-05-20.

\bibitem[{Frohmann et~al.(2024)Frohmann, Sterner, Vuli{\'c}, Minixhofer, and Schedl}]{frohmann2024segment}
Markus Frohmann, Igor Sterner, Ivan Vuli{\'c}, Benjamin Minixhofer, and Markus Schedl. 2024.
\newblock Segment any text: A universal approach for robust, efficient and adaptable sentence segmentation.
\newblock In \emph{Proceedings of the 2024 Conference on Empirical Methods in Natural Language Processing}, pages 11908--11941.

\bibitem[{Furman et~al.(2023)Furman, Torres, Rodr{\'\i}guez, Letzen, Martinez, and Alemany}]{furman-etal-2023-high}
Dami{\'a}n Furman, Pablo Torres, Jos{\'e} Rodr{\'\i}guez, Diego Letzen, Maria Martinez, and Laura Alemany. 2023.
\newblock \href {https://doi.org/10.18653/v1/2023.findings-emnlp.194} {High-quality argumentative information in low resources approaches improve counter-narrative generation}.
\newblock In \emph{Findings of the Association for Computational Linguistics: EMNLP 2023}, pages 2942--2956, Singapore. Association for Computational Linguistics.

\bibitem[{Gad-Elrab et~al.(2019)Gad-Elrab, Stepanova, Urbani, and Weikum}]{gad2019exfakt}
Mohamed~H Gad-Elrab, Daria Stepanova, Jacopo Urbani, and Gerhard Weikum. 2019.
\newblock Exfakt: A framework for explaining facts over knowledge graphs and text.
\newblock In \emph{Proceedings of the twelfth ACM international conference on web search and data mining}, pages 87--95.

\bibitem[{Grattafiori et~al.(2024)Grattafiori, Dubey, Jauhri, Pandey, Kadian, Al-Dahle, Letman, Mathur, Schelten, Vaughan et~al.}]{grattafiori2024llama}
Aaron Grattafiori, Abhimanyu Dubey, Abhinav Jauhri, Abhinav Pandey, Abhishek Kadian, Ahmad Al-Dahle, Aiesha Letman, Akhil Mathur, Alan Schelten, Alex Vaughan, and 1 others. 2024.
\newblock The llama 3 herd of models.
\newblock \emph{arXiv preprint arXiv:2407.21783}.

\bibitem[{Guo et~al.(2022)Guo, Schlichtkrull, and Vlachos}]{guo2022survey}
Zhijiang Guo, Michael Schlichtkrull, and Andreas Vlachos. 2022.
\newblock A survey on automated fact-checking.
\newblock \emph{Transactions of the Association for Computational Linguistics}, 10:178--206.

\bibitem[{Hameleers et~al.(2022)Hameleers, Van~der Meer, and Vliegenthart}]{hameleers2022civilized}
Michael Hameleers, Toni Van~der Meer, and Rens Vliegenthart. 2022.
\newblock Civilized truths, hateful lies? incivility and hate speech in false information--evidence from fact-checked statements in the us.
\newblock \emph{Information, Communication \& Society}, 25(11):1596--1613.

\bibitem[{He et~al.(2023)He, Ahamad, and Kumar}]{he2023reinforcement}
Bing He, Mustaque Ahamad, and Srijan Kumar. 2023.
\newblock Reinforcement learning-based counter-misinformation response generation: a case study of covid-19 vaccine misinformation.
\newblock In \emph{Proceedings of the ACM Web Conference 2023}, pages 2698--2709.

\bibitem[{Li et~al.(2023)Li, Liu, Xiao, and Shao}]{li2023making}
Chaofan Li, Zheng Liu, Shitao Xiao, and Yingxia Shao. 2023.
\newblock \href {https://arxiv.org/abs/2312.15503} {Making large language models a better foundation for dense retrieval}.
\newblock \emph{Preprint}, arXiv:2312.15503.

\bibitem[{Martone et~al.(2026)Martone, Bonaldi, and Guerini}]{martone2026assisted}
Genoveffa Martone, Helena Bonaldi, and Marco Guerini. 2026.
\newblock \href {https://arxiv.org/abs/2605.22435} {Assisted counterspeech writing at the crossroads of hate speech and misinformation}.
\newblock \emph{Preprint}, arXiv:2605.22435.

\bibitem[{Mathew et~al.(2019)Mathew, Saha, Tharad, Rajgaria, Singhania, Maity, Goyal, and Mukherjee}]{mathew2019thou}
Binny Mathew, Punyajoy Saha, Hardik Tharad, Subham Rajgaria, Prajwal Singhania, Suman~Kalyan Maity, Pawan Goyal, and Animesh Mukherjee. 2019.
\newblock \href {https://ojs.aaai.org/index.php/ICWSM/article/view/3237} {Thou shalt not hate: Countering online hate speech}.
\newblock In \emph{Proceedings of the International AAAI Conference on Web and Social Media}, volume~13, pages 369--380.

\bibitem[{Menini et~al.(2025)Menini, Russo, Aprosio, and Guerini}]{menini2025first}
Stefano Menini, Daniel Russo, Alessio~Palmero Aprosio, and Marco Guerini. 2025.
\newblock First-aid: the first annotation interface for grounded dialogues.
\newblock In \emph{Proceedings of the 63rd Annual Meeting of the Association for Computational Linguistics (Volume 3: System Demonstrations)}, pages 563--571.

\bibitem[{Mun et~al.(2023)Mun, Allaway, Yerukola, Vianna, Leslie, and Sap}]{munbeyond}
Jimin Mun, Emily Allaway, Akhila Yerukola, Laura Vianna, Sarah-Jane Leslie, and Maarten Sap. 2023.
\newblock Beyond denouncing hate: Strategies for countering implied biases and stereotypes in language.
\newblock In \emph{Proceedings of the 2023 Conference on Empirical Methods in Natural Language Processing}.

\bibitem[{Mun et~al.(2024)Mun, Buerger, Liang, Garland, and Sap}]{mun2024counterspeakers}
Jimin Mun, Cathy Buerger, Jenny~T Liang, Joshua Garland, and Maarten Sap. 2024.
\newblock Counterspeakers’ perspectives: Unveiling barriers and ai needs in the fight against online hate.
\newblock In \emph{Proceedings of the 2024 CHI Conference on Human Factors in Computing Systems}, pages 1--22.

\bibitem[{Myers~West(2018)}]{myers2018censored}
Sarah Myers~West. 2018.
\newblock Censored, suspended, shadowbanned: User interpretations of content moderation on social media platforms.
\newblock \emph{New Media \& Society}, 20(11):4366--4383.

\bibitem[{{OpenAI}(2024)}]{openai2024gpt4omini}
{OpenAI}. 2024.
\newblock {GPT-4o mini: advancing cost-efficient intelligence}.
\newblock \url{https://openai.com/index/gpt-4o-mini-advancing-cost-efficient-intelligence/}.
\newblock OpenAI blog post about the GPT-4o mini model release.

\bibitem[{{OpenAI}(2025)}]{openai_gpt41mini_2025}
{OpenAI}. 2025.
\newblock Gpt-4.1 mini.
\newblock \url{https://platform.openai.com/docs/models/gpt-4.1-mini}.
\newblock Accessed: 2026-05-25.

\bibitem[{Popat et~al.(2018)Popat, Mukherjee, Yates, and Weikum}]{popat2018declare}
Kashyap Popat, Subhabrata Mukherjee, Andrew Yates, and Gerhard Weikum. 2018.
\newblock \href {https://doi.org/10.18653/v1/D18-1003} {{D}e{C}lar{E}: Debunking fake news and false claims using evidence-aware deep learning}.
\newblock \emph{Proceedings of the 2018 Conference on Empirical Methods in Natural Language Processing}, pages 22--32.

\bibitem[{Press(2022)}]{apnews2022childrenquran}
The~Associated Press. 2022.
\newblock \href {https://apnews.com/article/fact-check-World-Cup-Qatar-Quran-314063624685} {Video of children reciting quran at qatar stadium is from 2021}.
\newblock Accessed: 2026-05-20.

\bibitem[{Robertson et~al.(1995)Robertson, Walker, Jones, Hancock-Beaulieu, Gatford et~al.}]{robertson1995okapi}
Stephen~E Robertson, Steve Walker, Susan Jones, Micheline~M Hancock-Beaulieu, Mike Gatford, and 1 others. 1995.
\newblock \emph{Okapi at TREC-3}.
\newblock British Library Research and Development Department.

\bibitem[{Russo(2025)}]{russo2025trenteam}
Daniel Russo. 2025.
\newblock Trenteam at multilingual counterspeech generation: Multilingual passage re-ranking approaches for knowledge-driven counterspeech generation against hate.
\newblock In \emph{Proceedings of the First Workshop on Multilingual Counterspeech Generation}, pages 77--91.

\bibitem[{Russo et~al.(2023{\natexlab{a}})Russo, Kaszefski-Yaschuk, Staiano, and Guerini}]{russo2023countering}
Daniel Russo, Shane Kaszefski-Yaschuk, Jacopo Staiano, and Marco Guerini. 2023{\natexlab{a}}.
\newblock \href {https://doi.org/10.18653/v1/2023.emnlp-main.703} {Countering misinformation via emotional response generation}.
\newblock \emph{Proceedings of the 2023 Conference on Empirical Methods in Natural Language Processing}, pages 11476--11492.

\bibitem[{Russo et~al.(2025{\natexlab{a}})Russo, Menini, Staiano, and Guerini}]{russo2025face}
Daniel Russo, Stefano Menini, Jacopo Staiano, and Marco Guerini. 2025{\natexlab{a}}.
\newblock Face the facts! evaluating rag-based pipelines for professional fact-checking.
\newblock In \emph{Proceedings of the 18th International Natural Language Generation Conference}, pages 846--865.

\bibitem[{Russo et~al.(2025{\natexlab{b}})Russo, Sadeghi, Menini, and Guerini}]{russo2025euroverdict}
Daniel Russo, Fariba Sadeghi, Stefano Menini, and Marco Guerini. 2025{\natexlab{b}}.
\newblock Euroverdict: A multilingual dataset for verdict generation against misinformation.
\newblock In \emph{Findings of the Association for Computational Linguistics: ACL 2025}, pages 16617--16634.

\bibitem[{Russo et~al.(2023{\natexlab{b}})Russo, Tekiroğlu, and Guerini}]{russo2023benchmarking}
Daniel Russo, Serra~Sinem Tekiroğlu, and Marco Guerini. 2023{\natexlab{b}}.
\newblock \href {https://doi.org/10.1162/tacl_a_00601} {{Benchmarking the Generation of Fact Checking Explanations}}.
\newblock \emph{Transactions of the Association for Computational Linguistics}, 11:1250--1264.

\bibitem[{Schieb and Preuss(2016)}]{schieb2016governing}
Carla Schieb and Mike Preuss. 2016.
\newblock Governing hate speech by means of counterspeech on facebook.
\newblock In \emph{66th ica annual conference, at fukuoka, japan}, pages 1--23.

\bibitem[{Sentence-Transformers(2021)}]{all-mpnet-base-v2}
Sentence-Transformers. 2021.
\newblock all-mpnet-base-v2.
\newblock \url{https://huggingface.co/sentence-transformers/all-mpnet-base-v2}.

\bibitem[{Snover et~al.(2006)Snover, Dorr, Schwartz, Micciulla, and Makhoul}]{snover2006study}
Matthew Snover, Bonnie Dorr, Richard Schwartz, Linnea Micciulla, and John Makhoul. 2006.
\newblock A study of translation edit rate with targeted human annotation.
\newblock In \emph{Proceedings of association for machine translation in the Americas}, volume 200, 6. Cambridge, MA.

\bibitem[{Stammbach and Ash(2020)}]{stammbach2020fever}
Dominik Stammbach and Elliott Ash. 2020.
\newblock e-fever: Explanations and summaries for automated fact checking.
\newblock \emph{Proceedings of the 2020 Truth and Trust Online (TTO 2020)}, pages 32--43.

\bibitem[{Team(2025)}]{qwen3technicalreport}
Qwen Team. 2025.
\newblock \href {https://arxiv.org/abs/2505.09388} {Qwen3 technical report}.
\newblock \emph{Preprint}, arXiv:2505.09388.

\bibitem[{Tekiro{\u{g}}lu et~al.(2020)Tekiro{\u{g}}lu, Chung, and Guerini}]{tekiroglu2020generating}
Serra~Sinem Tekiro{\u{g}}lu, Yi-Ling Chung, and Marco Guerini. 2020.
\newblock \href {https://doi.org/10.18653/v1/2020.acl-main.110} {Generating counter narratives against online hate speech: Data and strategies}.
\newblock \emph{Proceedings of the 58th Annual Meeting of the Association for Computational Linguistics}, pages 1177--1190.

\bibitem[{Turchi et~al.(2013)Turchi, Negri, and Federico}]{turchi2013coping}
Marco Turchi, Matteo Negri, and Marcello Federico. 2013.
\newblock Coping with the subjectivity of human judgements in mt quality estimation.
\newblock In \emph{Proceedings of the Eighth Workshop on Statistical Machine Translation}, pages 240--251.

\bibitem[{Vallecillo-Rodr{\'\i}guez et~al.(2023)Vallecillo-Rodr{\'\i}guez, Montejo-Ra{\'e}z, and Mart{\'\i}n-Valdivia}]{vallecillo2023automatic}
Maria~Estrella Vallecillo-Rodr{\'\i}guez, Arturo Montejo-Ra{\'e}z, and Maria~Teresa Mart{\'\i}n-Valdivia. 2023.
\newblock Automatic counter-narrative generation for hate speech in spanish.
\newblock \emph{Procesamiento del lenguaje natural}, 71:227--245.

\bibitem[{Vidgen and Derczynski(2020)}]{vidgen2020directions}
Bertie Vidgen and Leon Derczynski. 2020.
\newblock Directions in abusive language training data, a systematic review: Garbage in, garbage out.
\newblock \emph{Plos one}, 15(12):e0243300.

\bibitem[{Wardle(2024)}]{wardle2024conceptual}
Claire Wardle. 2024.
\newblock A conceptual analysis of the overlaps and differences between hate speech, misinformation and disinformation.
\newblock \emph{Department of Peace Operations (DPO). Office of the Special Adviser on the Prevention of Genocide (OSAPG). United Nations}.

\bibitem[{Wintersieck(2017)}]{wintersieck2017debating}
Amanda~L Wintersieck. 2017.
\newblock Debating the truth: The impact of fact-checking during electoral debates.
\newblock \emph{American politics research}, 45(2):304--331.

\bibitem[{Wu et~al.(2022)Wu, Luan, Rashkin, Reitter, Hajishirzi, Ostendorf, and Tomar}]{wu2022conqrr}
Zeqiu Wu, Yi~Luan, Hannah Rashkin, David Reitter, Hannaneh Hajishirzi, Mari Ostendorf, and Gaurav~Singh Tomar. 2022.
\newblock Conqrr: Conversational query rewriting for retrieval with reinforcement learning.
\newblock In \emph{Proceedings of the 2022 Conference on Empirical Methods in Natural Language Processing}, pages 10000--10014.

\bibitem[{Ye et~al.(2023)Ye, Fang, Li, and Yilmaz}]{ye2023enhancing}
Fanghua Ye, Meng Fang, Shenghui Li, and Emine Yilmaz. 2023.
\newblock Enhancing conversational search: Large language model-aided informative query rewriting.
\newblock In \emph{Findings of the Association for Computational Linguistics: EMNLP 2023}, pages 5985--6006.

\bibitem[{Zeng and Gao(2024)}]{zeng-gao-2024-justilm}
Fengzhu Zeng and Wei Gao. 2024.
\newblock \href {https://doi.org/10.1162/tacl_a_00649} {{J}usti{LM}: Few-shot justification generation for explainable fact-checking of real-world claims}.
\newblock \emph{Transactions of the Association for Computational Linguistics}, 12:334--354.

\bibitem[{Zhang et~al.(2019)Zhang, Kishore, Wu, Weinberger, and Artzi}]{zhang2019bertscore}
Tianyi Zhang, Varsha Kishore, Felix Wu, Kilian~Q Weinberger, and Yoav Artzi. 2019.
\newblock Bertscore: Evaluating text generation with {BERT}.
\newblock \emph{arXiv preprint arXiv:1904.09675}.

\bibitem[{Zhang et~al.(2025)Zhang, Li, Long, Zhang, Lin, Yang, Xie, Yang, Liu, Lin, Huang, and Zhou}]{qwen3embedding}
Yanzhao Zhang, Mingxin Li, Dingkun Long, Xin Zhang, Huan Lin, Baosong Yang, Pengjun Xie, An~Yang, Dayiheng Liu, Junyang Lin, Fei Huang, and Jingren Zhou. 2025.
\newblock Qwen3 embedding: Advancing text embedding and reranking through foundation models.
\newblock \emph{arXiv preprint arXiv:2506.05176}.

\end{thebibliography}

\clearpage
\appendix

\section{Appendix}

\subsection{Document collection} \label{appendix:doc_collection}
\begin{figure*}
    \centering
    \includegraphics[width=\textwidth]{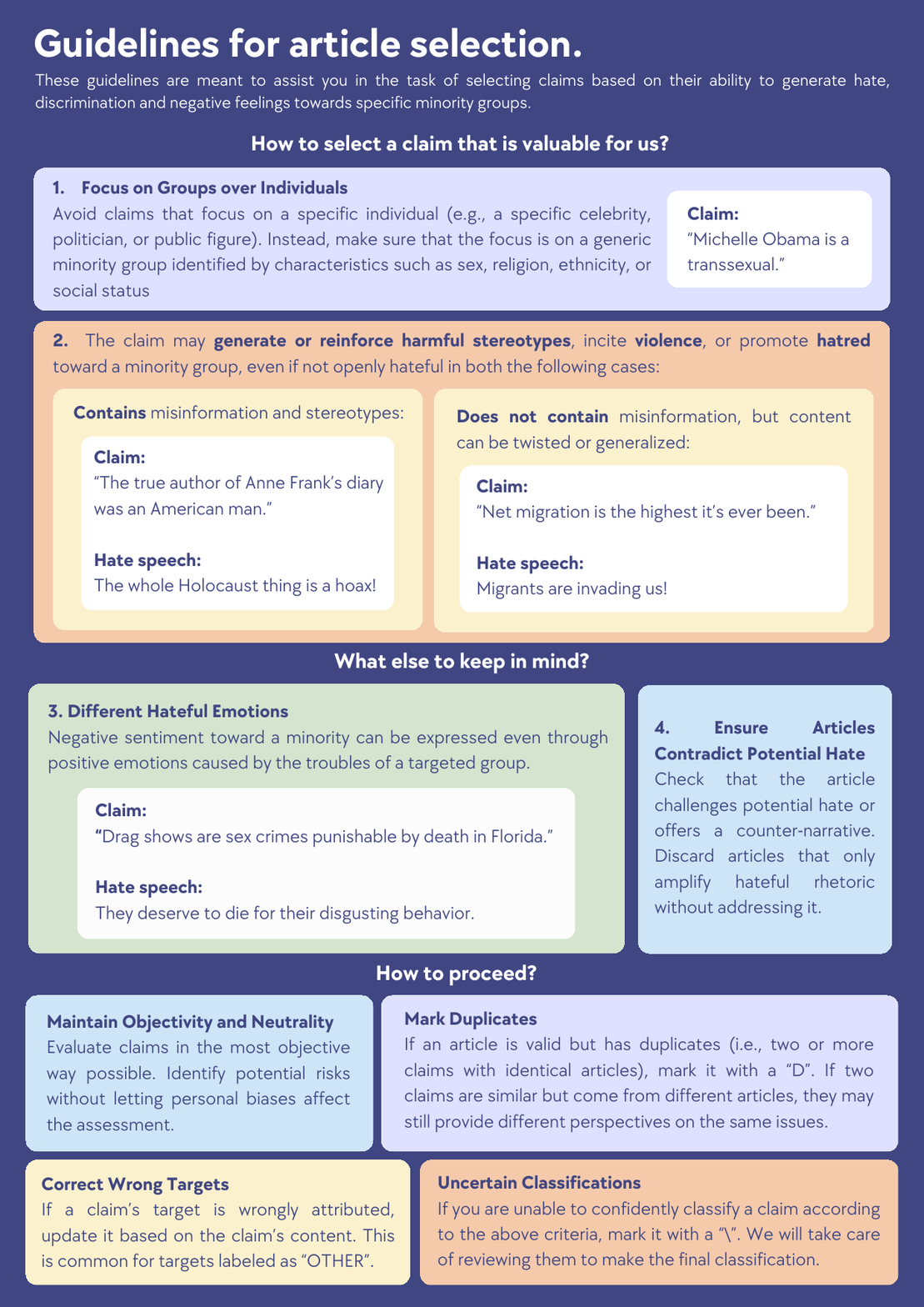}
    \caption{Guidelines for selecting articles with potential to fuel hate and discrimination.}
    \label{fig:claim_guide}
\end{figure*}

\paragraph{Fact-checking articles} 
To retrieve multilingual fact-checking articles from Google Fact Check Tools Explorer, the English keyword dictionary produced by \citet{martone2026assisted} was automatically translated into Polish, Spanish, and Italian using \texttt{deepl}.\footnote{\url{https://pypi.org/project/deepl/}} The translated keywords were then submitted to our NGO collaborators, who worked on the lists in their respective native languages. They contributed by fixing potentially wrong translations, removing words that were not used in their linguistic context, and adding relevant synonyms. The complete keyword lists obtained from this process are available in Table \ref{tab:keywords}. Additionally, Figure \ref{fig:claim_guide} shows the guidelines used to select the fact-checking articles to keep as external knowledge, based on their ability to create or fuel hate against the considered target groups. Below we provide an example by The Associated \citet{apnews2022childrenquran}:

\begin{fcbox}
\small
\noindent
\textbf{Claim}: A video shows the 2022 FIFA World Cup in Qatar opening with children reciting the Quran. \\

\noindent \textbf{Fact-checking}: False. The video was filmed on Oct. 22, 2021, and shows an inauguration ceremony for the Al Thumama Stadium, a World Cup venue in Doha. THE FACTS: The World Cup began Sunday in Qatar, the first Arab or Muslim nation to host the competition. Social media users shared a year-old video from the inauguration of the stadium, falsely claiming it showed the tournament’s opening ceremony[...].

\end{fcbox}

\begin{table*}
    \centering
    \small
    \begin{tabularx}{\textwidth}{lXXXX}
    \toprule
         \multicolumn{1}{c}{\textbf{Target}} & \multicolumn{4}{c}{\textbf{Keywords}} \\ 
         & \multicolumn{1}{c}{\textbf{English}} 
         & \multicolumn{1}{c}{\textbf{Italian}} 
         & \multicolumn{1}{c}{\textbf{Spanish}}
         & \multicolumn{1}{c}{\textbf{Polish}} \\
         \midrule
         Muslims&  muslim, islam, terrorist, jihadi, jihad, ragheadterror, arab, koran, quran, sharia, towel head, rag head& musulmano, islam, terrorist, jihadista, jihad, arabo, giornale, corano, sharia & musulmán, islam, terrorista, yihadista, yihad, árabe, corán, sharia & szmatogłowy, muzułmanin, islam, terrorysta, dżihadysta, dżihad, Arab, szariat\\
         LGBTQIA+&  gay, homosexual, homosexuality, lgbt, lgbt+, lgbti, lgbtq+, lgbtq, faggot, gender, lesbian, trans, transgender, transsexual, queer, sexual, sex, heterosexual, dyke, gay pride & trans, frocio, lgbtq, lgbtq+, genere, lesbica, transgender, lgbt+, queer, sessuale, sesso, lgbti, eterosessuale, transessuale, lgbt, gay pride, omosessuale, gay, omosessualità & trans, maricón, lgbtq, lgbtq+, género, lesbiana, transgénero, lgbt+, queer, sexual, sexo, lgbti, heterosexual, bollera, transexual, lgbt, orgullo gay, homosexual, gay, homosexualidad & trans, pedał, ciota, lgbtq, lgbtq+, płeć, lesbijka, transpłciowy, lgbt+, queer, seksualny, płeć biologiczna, heteroseksualny, lesba, transseksualista, transseksualny, marsz równości, homoseksualista, homoseksualny, gej, gejowski, homoseksualność\\
         Migrants&  migrant, immigrant, refugee, immigration, foreigner, migration, foreign, rapefugees, invasion, invade, refugeesnotwelcome & rifugiato, i rifugiati non sono i benvenuti, invadere, invasione, rapefugees, estero, migrazione, straniero, immigrazione, immigrato, migrante & refugiado, invadir, invasión, violadores, extranjero, migración, inmigración, inmigrante, migrante & uchodźca, uchodźcy nie są mile widziani, najeżdżać, inwazja, rapefugees, obcy, migracja, cudzoziemiec, imigracja, imigrant, migrant, refugeesnotwelcome\\
         Women&  woman, feminism, feminist, gender, female, harassment, feminazi, shithole, cunt, blameonenotall, notallmen, victimcard, sexual assault, victim card & violenza sessuale, carta della vittima, molestie, colpa di non tutti, fica, cesso, feminazi, femminile, genere, femminista, femminismo, donna, scheda-vittima, non tutti gli uomini & agresión sexual, acoso, feminazi, mujer, género, feminista, feminismo & napaść na tle seksualnym, karta ofiary, nękanie, notallmen, blameonenotall, pizda, zadupie, feminazistka, kobieta, płeć, feministka, feminizm, kurwa, feministyczny\\
         People with\\
         disabilities&  disabled, disability, autistic, blind, deaf, retard, downies, downy, paralympics, wheelchair& disabile, disabilità, autistico, cieco, sordo, ritardato, down, paralimpiadi, sedia a rotelle & discapacitado, discapacidad, autista, ciego, sordo, retrasado, down, paralímpicos, silla de ruedas & niepełnosprawny, niepełnosprawność, autystyczny, niewidomy, głuchy, niedorozwinięty\\
         Jews&  jew, jewish, holocaust, judaism, nazi, nazism, genocide& ebreo, ebraico, olocausto, nazism, nazi, genocidio, ebraismo & judío, holocausto, nazismo, nazi, genocidio, judaísmo & Żyd, żydowski, holokaust, nazizm, nazista, nazistowski, ludobójstwo, judaizm\\ \bottomrule
    \end{tabularx}
    \caption{Multilingual keywords used to query Google Fact Check Tools for retrieving fact-checking articles related to the groups of our interest.}
    \label{tab:keywords}
\end{table*}

\paragraph{NGO reports} The URL domains from which we extracted the NGO reports in Italian, Spanish, and Polish are the following: \url{https://biblioteka.ceo.org.pl}, \url{https://humanityinaction.org/country/poland/}, \url{https://fundacionadecco.org}, \url{https://holocaustcentrenorth.org.uk}, \url{https://maldita.es}, \url{https://migrantesenigualdad.es}, \url{https://oko.press}, \url{https://porcausa.org}, \url{https://uchodzcy.info}, \url{https://uprzedzuprzedzenia.org}, \url{https://www.facta.news}, \url{https://www.huffpost.com}, \url{https://www.poradnikzdrowie.pl}.
Below is an example in Italian by \citet{facta2022LBGTmyths} in Facta:

\begin{ngobox}
\small
\noindent
\textbf{Myth}: Le persone trans sono uomini vestiti da donne (o viceversa). \\

\noindent \textbf{Anti-stereotype}: Gli uomini vestiti da donne e le donne vestite da uomini sono chiamati “crossdresser” e il loro comportamento non ha necessariamente a che fare con l’identità di genere percepita o l’orientamento sessuale. [...]

\end{ngobox}

\subsection{Human-Machine Collaboration Details} \label{appendix:collaboration_strategies}
\paragraph{Pre-compiled Strategy} For generation we use \texttt{gpt-4o-mini-2024-07-18} (Temp: 0.7, Max Tokens: 500). Documents are split with SaT (\texttt{sat-1l-sm}, universal dependencies format). We use the following prompt:

\begin{lstlisting}
Given the following article, generate a dialogue in <LANGUAGE> between a person spreading hate against <TARGET> and an NGO operator who provides polite and informed counterstatements based on the article. The hater does not give up easily on their opinions. The dialogue must include at most <TURNS_NUMBER> exchanges. Return the response in JSON format where each turn is clearly marked by the speaker. Use the following format: { "dialogue": [ {"speaker": "<SPEAKER_1>", "text": "[Dialogue]"}, {"speaker": "<SPEAKER_2>", "text": "[Dialogue]"}, ... ] } Ensure the structure remains consistent throughout.
\end{lstlisting}

\subsubsection*{Interactive Strategy} 

\begin{itemize}
    \item \textbf{Generators:} We run preliminary experiments with \texttt{gpt-4o-mini} (same hyperparameters as pre-compiled), but we encounter issues as the model refuses to generate the hateful/misinformed turns. Therefore, we fine-tune \texttt{Llama-3.1-8B} via QLoRA \cite{dettmers2023qlora} (4-bit, LoRA $r=32$, $\alpha=64$, drop=0), on an initial corpus of 1108 dialogues collected at the moment with the Manual, Pre-compiled and Translation strategies. Such data is split in the training and validation set with a 80-20 proportion (886 dialogues for training and 222 for validation). The dataset is split by ensuring that there is not a dialogue on the same article which is present both in the training and validation set. Moreover, we stratify the data according to the targeted minority in the dialogue and the language, so that there is a similar distribution in the two subsets according to these variables. Training runs for 5 epochs, max sequence length 3,000, using AdamW (LR: $5\times10^{-5}$, cosine decay, 3\% warmup, weight decay 0.01) with an effective batch size of 18 and early stopping (patience=10, eval every 10 steps). The prompts employed for training are the same used to collect dialogues with the interactive strategy, and is shown below.
    \item \textbf{Retrieval:} We test both \texttt{BAAI/bge-m3} and a fine-tuned \texttt{BAAI/bge-reranker-v2-m3}.

    Also for preparing the training data for the reranker we employed the portion of dialogues collected at the moment with the Manual, Pre-compiled and Translation strategies. To prepare the data, each article document was segmented into sentences using the spaCy English sentencizer. For every dialogue, hater's turns were treated as queries, while the subsequent counterspeech turns provided grounded evidence spans annotated in the data. These evidence spans were also sentence-segmented and used as positive passages. Negative passages were obtained by extracting all remaining article sentences that did not overlap with the normalized positive spans, where normalization included lowercasing, whitespace stripping, and Unicode normalization to remove accent variations. Each resulting training instance therefore consisted of a query, a set of positive evidence sentences, and a set of negative sentences sampled from the same source article. Instances without negative examples were removed, thus obtaining 2659 training examples.
    The reranker was fine-tuned for 3 epochs using contrastive learning (no distillation, max len 512, FP16, effective batch size 8, negative group size 8). Optimized with AdamW (LR: $6\times10^{-5}$, 10\% warmup, weight decay 0.01, gradient checkpointing).
\end{itemize}

\subsubsection*{Prompt for Interactive Strategy}
\begin{itemize}
    \item First Turn:
\begin{lstlisting}
Given the provided claim and fact-checking article, pretend to be a hateful person and provide a hypothetical statement in <LANGUAGE> spreading harmful or discriminatory rhetoric against <TARGET> based on the claim (contained in <claim> tags) and the article (contained in <article> tags). The information needed to answer your message must be present exclusively in the article.
<article><ARTICLE></article>
<claim><CLAIM></claim>
\end{lstlisting}

    \item Hater:
\begin{lstlisting} 
Given the dialogue history and the latest turn which references context in the dialogue history, formulate a reply to the last turn in <LANGUAGE> pretending to be a person who is spreading hate against <TARGET>. Do not repeat prior turns. The information needed to answer your message must be present exclusively in the provided article (contained in <article> tags).
<article><ARTICLE></article>
\end{lstlisting}

    \item Operator:
\begin{lstlisting}
You are an operator who provides polite and informed counterstatements in <LANGUAGE> to hateful and misinformed messages. Your task is to provide a relevant answer to the hater using the provided evidence and past dialogue history. The evidence is contained in <context> tags. Answer by referring exclusively to the content of the document.
<context><GROUND_TEXT></context>
\end{lstlisting}
\end{itemize}

\noindent \textbf{Translation Strategy}: We use \texttt{facebook/hf-seamless-m4t-large} (\texttt{max\_length}=512, \texttt{do\_sample=False}).

\subsection{More details on dataset description} \label{appendix:data_description}
Table \ref{tab:target_coverage} reports the distribution of dialogues according to target, Table \ref{tab:lang_coverage} according to language and Table \ref{tab:target_language_coverage} according to both target and language. ``Multitarget'' refers to dialogues that can be referred to more than one target at once.

\begin{table}
\small
\centering
\begin{tabular}{lrr}
\toprule
\textbf{TARGET} & \textbf{Tot} & \textbf{\%} \\
\midrule
DISABLED & 141 & 7.00 \\
JEWS & 44 & 2.18 \\
LGBTQIA+ & 264 & 13.10 \\
MIGRANTS & 615 & 30.52 \\
MUSLIMS & 376 & 18.66 \\
POC & 225 & 11.17 \\
WOMEN & 336 & 16.67 \\
multitarget & 11 & 0.55 \\
\bottomrule
\end{tabular}
\caption{Distribution of dialogues according to target.}
\label{tab:target_coverage}
\end{table}

\begin{table}
\small
\centering
\resizebox{\columnwidth}{!}{%
\begin{tabular}{ll|rrr|rrr}
\toprule
\textbf{Docs} & \textbf{Lang.} & \textbf{Mono.} & \textbf{Trans.} & \textbf{Tot.} & \textbf{\% Mono.} & \textbf{\% Trans.} & \textbf{\% Tot.} \\
\midrule
\multirow{5}{*}{1} & EN & 336 & 0 & 336 & 16.67 & 0.00 & 16.67 \\
& ES & 110 & 194 & 304 & 5.46 & 9.63 & 15.09 \\
& IT & 108 & 194 & 302 & 5.36 & 9.63 & 14.99 \\
& MT & 37 & 256 & 293 & 1.84 & 12.70 & 14.54 \\
& PL & 137 & 193 & 330 & 6.80 & 9.58 & 16.38 \\
\midrule
\multirow{5}{*}{2} & EN & 50 & 0 & 50 & 2.48 & 0.00 & 2.48 \\
& ES & 50 & 50 & 100 & 2.48 & 2.48 & 4.96 \\
& IT & 50 & 50 & 100 & 2.48 & 2.48 & 4.96 \\
& MT & 50 & 50 & 100 & 2.48 & 2.48 & 4.96 \\
& PL & 50 & 50 & 100 & 2.48 & 2.48 & 4.96 \\
\bottomrule
\end{tabular}
}
\caption{Language distribution across one and two-document dialogues.}
\label{tab:lang_coverage}
\end{table}

\begin{table}
\small
\centering
\begin{tabular}{llrr}
\toprule
\textbf{TARGET} & \textbf{language} & \textbf{Tot.} & \textbf{\%} \\
\midrule
\multirow{5}{*}{DISABLED} & EN & 26 & 1.29 \\
 & ES & 21 & 1.04 \\
 & IT & 27 & 1.34 \\
 & MT & 34 & 1.69 \\
 & PL & 33 & 1.64 \\
 \midrule
\multirow{5}{*}{JEWS}  & EN & 10 & 0.50 \\
 & ES & 5 & 0.25 \\
 & IT & 5 & 0.25 \\
 & MT & 17 & 0.84 \\
 & PL & 7 & 0.35 \\
  \midrule
\multirow{5}{*}{LGBT+}  & EN & 51 & 2.53 \\
 & ES & 66 & 3.28 \\
 & IT & 50 & 2.48 \\
 & MT & 44 & 2.18 \\
 & PL & 53 & 2.63 \\
  \midrule
\multirow{5}{*}{MIGRANTS}  & EN & 102 & 5.06 \\
 & ES & 141 & 7.00 \\
 & IT & 133 & 6.60 \\
 & MT & 95 & 4.71 \\
 & PL & 144 & 7.15 \\
  \midrule
\multirow{5}{*}{MUSLIMS}  & EN & 81 & 4.02 \\
 & ES & 65 & 3.23 \\
 & IT & 69 & 3.42 \\
 & MT & 84 & 4.17 \\
 & PL & 77 & 3.82 \\
  \midrule
\multirow{5}{*}{POC}  & EN & 46 & 2.28 \\
 & ES & 44 & 2.18 \\
 & IT & 43 & 2.13 \\
 & MT & 51 & 2.53 \\
 & PL & 41 & 2.03 \\
  \midrule
ROMANI & IT & 3 & 0.15 \\
 \midrule
\multirow{5}{*}{WOMEN}  & EN & 67 & 3.33 \\
 & ES & 60 & 2.98 \\
 & IT & 70 & 3.47 \\
 & MT & 66 & 3.28 \\
 & PL & 73 & 3.62 \\
  \midrule
\multirow{5}{*}{multitarget}  & EN & 3 & 0.15 \\
 & ES & 2 & 0.10 \\
 & IT & 2 & 0.10 \\
 & MT & 2 & 0.10 \\
 & PL & 2 & 0.10 \\
\bottomrule
\end{tabular}
\caption{Distribution of dialogues according to target and language: ``multitarget'' refers to dialogues that can be referred to more than one target at once.}
\label{tab:target_language_coverage}
\end{table}

\subsection{More analyses: monolingual data} \label{app:more_monolingual_analyses}

\begin{table}
\small
\centering
\resizebox{\columnwidth}{!}{%
\begin{tabular}{llllllll}
\toprule
\textbf{Docs} & \textbf{Strat.} & \textbf{HTER} & \textbf{Time} & \textbf{Ground} & \textbf{RR$_{or}$} & \textbf{RR$_{ed}$} & \textbf{RR $\Delta$} \\
\midrule
\multirow{3}{*}{1}  & Interactive & 0.408 & 169.544 & 29.532 & 8.775 & 7.676 & {\ul -1.099} \\
& Manual & - & 179.446 & 27.558 & - & \textbf{2.396} & - \\
& Pre-compiled & {\ul 0.314} & {\ul 105.374} & {\ul 38.926} & \textbf{4.047} & 3.994 & -0.053 \\
\midrule
\multirow{3}{*}{2}  & Interactive & 0.434 & 159.962 & 33.81 & 12.198 & 8.219 & \textbf{-3.979} \\
& Manual & - & 515.89 & 30.62 & - & {\ul 3.237} & - \\
& Pre-compiled & \textbf{0.302} & \textbf{85.388} & \textbf{60.180} & {\ul 5.725} & 4.686 & -1.039 \\
\bottomrule
\end{tabular}
}
\caption{Annotation effort results according to number of documents and strategy.}
\label{tab:docs_strategy_annot_effort}
\end{table}

Table \ref{tab:docs_strategy_annot_effort} shows the annotation effort results obtained by jointly considering the number of documents and strategy. The pre-compiled strategy is the one with the lowest HTER, longest ground, and shortest annotation time in both one- and two-documents strategies.
Manually writing dialogues requires the highest annotation time in both document configurations, but it also guarantees the lowest RR$_{ed}$: in settings where LLM outputs cannot be modified, manual annotation remains the best source of high-quality data.
consistently with the results grouped by strategy only, 
the interactive strategy produces the most repetitive generations in both document settings, as shown by the RR$_{or}$, while also obtaining the greatest reduction after post-editing. Syntactic metrics results are coherent with those shown by strategy only.

\begin{table*}
\small
\centering
\resizebox{\textwidth}{!}{%
\begin{tabular}{lrrrrrrrrrrrrr}
\toprule
\textbf{Docs} & \textbf{Strat.} & \textbf{MSD$_{or}$} & \textbf{MSD$_{ed}$} & \textbf{$\text{MSD}_{\Delta}$} & \textbf{ASD$_{or}$} & \textbf{ASD$_{ed}$} & \textbf{$\text{ASD}_{\Delta}$} & \textbf{NST$_{or}$} & \textbf{NST$_{ed}$} & \textbf{$\text{NST}_{\Delta}$} & \textbf{CW$_{or}$} & \textbf{CW$_{ed}$} & \textbf{$\text{CW}_{\Delta}$} \\
\midrule
\multirow{3}{*}{1}  & Interactive & \textbf{5.240} & 4.874 & -0.366 & \textbf{4.373} & 4.089 & -0.284 & {\ul 2.024} & 2.006 & -0.018 & 0.323 & 0.314 & -0.009 \\
& Manual & - & {\ul 5.033} & - & - & {\ul 4.132} & - & - & 2.194 & - & - & 0.275 & - \\
& Pre-compiled & 4.676 & 4.86 & \textbf{0.184} & 3.838 & 3.913 & \textbf{0.075} & 1.984 & 2.217 & \textbf{0.233} & 0.288 & 0.286 & {\ul -0.002} \\
\midrule
\multirow{3}{*}{2}  & Interactive & {\ul 4.681} & 4.685 & 0.004 & {\ul 3.888} & 3.917 & {\ul 0.029} & 1.936 & 2.01 & 0.074 & \textbf{0.358} & \textbf{0.348} & -0.01 \\
& Manual & - & \textbf{5.609} & - & - & \textbf{4.339} & - & - & \textbf{2.405} & - & - & 0.132 & - \\
& Pre-compiled & 4.52 & 4.558 & {\ul 0.038} & 3.66 & 3.664 & 0.004 & \textbf{2.206} & {\ul 2.343} & {\ul 0.137} & {\ul 0.334} & {\ul 0.336} & \textbf{0.002} \\
\bottomrule
\end{tabular}
}
\caption{Syntactic metrics results according to number of documents and strategy.}
\label{tab:docs_strategy_syntactic}
\end{table*}

\subsection{More analyses: translated data} \label{appendix:translation_analyses}
Table \ref{tab:trans_docs_lang_annot_effort} and \ref{tab:trans_docs_lang_syntactic} report, respectively, the annotation effort and syntactic metrics results for translated dialogues according to the number of documents and language. Results are consistent with those obtained when grouping by language only, as discussed in \S\ref{subsec:translated_data}.

\begin{table}
\small
\centering
\resizebox{\columnwidth}{!}{
\begin{tabular}{llrrrrrr}
\toprule
\textbf{Docs} & \textbf{Lang.} & \textbf{HTER} & \textbf{Time} & \textbf{Ground} & \textbf{RR$_{or}$} & \textbf{RR$_{ed}$} & \textbf{RR $\Delta$} \\
\midrule
\multirow{4}{*}{1}  & ES & \textbf{1.010} & {\ul 41.15} & 17.43 & 3.115 & 4.058 & 0.943 \\
& IT & {\ul 1.077} & \textbf{33.37} & {\ul 17.54} & 2.564 & 3.076 & 0.512 \\
& MT & 1.404 & 153.73 & \textbf{18.04} & \textbf{2.162} & \textbf{2.135} & {\ul -0.027} \\
& PL & 1.287 & 94.49 & {\ul 17.54} & {\ul 2.262} & {\ul 2.165} & \textbf{-0.097} \\
\midrule
\multirow{4}{*}{2}  & ES & {\ul 1.019} & {\ul 129.55} & {\ul 30.88} & 3.166 & 4.361 & 1.195 \\
& IT & \textbf{1.001} & \textbf{28.22} & 30.62 & 2.844 & 3.843 & 0.999 \\
& MT & 1.439 & 206.26 & 30.62 & {\ul 2.149} & {\ul 2.160} & {\ul 0.011} \\
& PL & 1.356 & 173.06 & \textbf{31.35} & \textbf{2.081} & \textbf{1.783} & \textbf{-0.298} \\
\bottomrule
\end{tabular}
}
\caption{Annotation effort results for translated dialogues according to the number of documents and language.}
\label{tab:trans_docs_lang_annot_effort}
\end{table}

\begin{table*}
\small
\centering
\resizebox{\textwidth}{!}{
\begin{tabular}{llrrrrrrrrrrrr}
\toprule
\textbf{Docs} & \textbf{Lang.} & \textbf{MSD$_{or}$} & \textbf{MSD$_{ed}$} & \textbf{$\text{MSD}_{\Delta}$} & \textbf{ASD$_{or}$} & \textbf{ASD$_{ed}$} & \textbf{$\text{ASD}_{\Delta}$} & \textbf{NST$_{or}$} & \textbf{NST$_{ed}$} & \textbf{$\text{NST}_{\Delta}$} & \textbf{CW$_{or}$} & \textbf{CW$_{ed}$} & \textbf{$\text{CW}_{\Delta}$} \\
\midrule
\multirow{4}{*}{1} & ES & {\ul 4.622} & 4.630 & 0.008 & 3.966 & 3.911 & {\ul -0.055} & 1.881 & 1.949 & 0.068 & 0.309 & 0.307 & -0.002 \\
& IT & \textbf{4.658} & {\ul 4.655} & -0.003 & 3.874 & 3.804 & -0.07 & 1.979 & 2.049 & 0.07 & 0.357 & 0.356 & -0.001 \\
& MT & - & - & - & - & - & - & - & - & - & 0.252 & 0.255 & \textbf{0.003} \\
& PL & 4.621 & \textbf{4.719} & 0.098 & \textbf{4.091} & 3.897 & -0.194 & 1.726 & 2.081 & \textbf{0.355} & {\ul 0.415} & {\ul 0.409} & -0.006 \\
\midrule
\multirow{4}{*}{2} & ES & {\ul 4.911} & {\ul 5.010} & {\ul 0.099} & 3.983 & \textbf{3.985} & \textbf{0.002} & {\ul 2.207} & 2.276 & 0.069 & 0.309 & 0.308 & -0.001 \\
& IT & \textbf{5.036} & 5.000 & -0.036 & 3.952 & 3.868 & -0.084 & \textbf{2.365} & \textbf{2.398} & 0.033 & 0.365 & 0.364 & -0.001 \\
& MT & - & - & - & - & - & - & - & - & - & 0.253 & 0.252 & {\ul -0.001} \\
& PL & 4.878 & \textbf{5.013} & \textbf{0.135} & {\ul 4.077} & {\ul 3.959} & -0.118 & 2.023 & {\ul 2.355} & {\ul 0.332} & \textbf{0.428} & \textbf{0.425} & -0.003 \\
\bottomrule
\end{tabular}
}
\caption{Syntactic metrics results for translated dialogues according to the number of documents and language.}
\label{tab:trans_docs_lang_syntactic}
\end{table*}

\subsection{Retrieval experiment details} \label{appendix:retrieval_details}
We chunk documents with Llamaindex (chunk\_size=256, chunk\_overlap=64). To obtain positive chunks for testing, we automatically align the human-annotated \textit{ground text} spans to the obtained document chunks. 
For each query (i.e. hate speech turn), the search space is restricted to chunks belonging only to the annotated target documents via a document-to-chunk index mapping. Each ground text is then compared against all candidate chunks using a normalized longest-common-substring overlap score computed with \texttt{SequenceMatcher}. The 2 chunks with highest overlap over a 0.5 fixed threshold are labeled as positive examples. This produces query-level sets of positive chunk indices aligned directly to the pre-computed chunk corpus. After matching, we remove any duplicates to ensure that each chunk is mapped at most once to each query. Retrieval experiments took roughly 1 hour to run on a NVIDIA A40, Ampere GPU.

We use the \texttt{sentence-transformers}\footnote{\url{https://www.sbert.net/}} library for both BGE-M3 and Qwen3-Embedding. For what regards the latter, we also make sure to append the \textit{<|endoftext|>} token to each instruction, as suggested by the authors:
\texttt{{Instruction} {Query}<|endoftext|>}. For Qwen3, we employ the following instruction: \texttt{Given a query, retrieve relevant passages that refute the query}.

For both monolingual and cross-lingual setup, we employ as queries all hater's turns followed by a counterspeech grounded on external knowledge (we discard counterspeech examples not grounded on any external knowledge such as clarifying questions). In this way, we obtain a set of 2710 queries for the cross-lingual setup and 2409 queries for the monolingual setup.

For the rewritten query configuration, query is rewritten with GPT-5.4, max\_token = 1024, n = 1, stop = None, temperature = 0, using the following prompt from \citet{ye2023enhancing}:
\begin{itemize}
    \item System prompt: 
    \begin{lstlisting}
    "Given a query and its context, rewrite the query in {language} and decontextualize it by addressing coreference and omission issues. The resulting query should retain its original meaning and be as informative as possible, and should not duplicate any previous query in the context."
    \end{lstlisting}
    \item User prompt:
    \begin{lstlisting}
    "Context: {context}\nQuery: {question}\nRewrite: "
    \end{lstlisting}
\end{itemize}

\paragraph{Retrieval Cross-lingual results} Table \ref{tab:retr_crosslingual} presents the cross-lingual results of the retrieval experiment. We include BM25 in the cross-lingual evaluation purely as a lower-bound baseline to quantify the lexical overlap (e.g., via shared entities, proper nouns, and loanwords) between the non-English queries and English documents. As expected, BM25 experiences a near-total collapse, confirming that surface-level keyword matching is fundamentally inadequate for this task and validating the necessity of dense cross-lingual embedding spaces.

As for the monolingual setting, Qwen3 performs best across all query configurations, followed by BGE-M3. Interestingly, in the cross-lingual configuration both dense embedding models outperform their monolingual counterparts. 
We hypothesize that, 
because these embedders are predominantly optimized on high-quality English corpora, 
mapping a non-English query directly into this highly refined English space yields cleaner alignment and stronger retrieval signals than navigating the noisier, lower-resource local language document spaces inherent to the monolingual task. 
Finally, differently from the monolingual setting, where $Q_{\text{DC}}$ is almost always the best performing query formulation, in the cross-lingual scenario for both BM25 and BGE-M3 $Q_{\text{R}}$ outperforms $Q_{\text{DC}}$.

\begin{table}[ht]
\small
\centering
\resizebox{\columnwidth}{!}{
\begin{tabular}{llrrr}
\toprule
\textbf{Model} & \textbf{Query} & \textbf{Hit@10} & \textbf{MAP@10} & \textbf{Recall@10} \\
\midrule
\multirow{3}{*}{BM25} 
 & Q        & 0.034 & 0.015 & 0.029 \\
 & Q$_{\text{DC}}$ & 0.058 & 0.023 & 0.050 \\
 & Q$_{\text{R}}$  & \textbf{0.063} & \textbf{0.029} & \textbf{0.054} \\
\midrule
\multirow{3}{*}{BGE-M3} 
 & Q        & 0.432 & 0.225 & 0.389 \\
 & Q$_{\text{DC}}$ & 0.649 & 0.337 & 0.601 \\
 & Q$_{\text{R}}$  & \textbf{0.660} & \textbf{0.361} & \textbf{0.611} \\
\midrule
\multirow{3}{*}{Qwen3} 
 & Q        & 0.457 & 0.242 & 0.411 \\
 & Q$_{\text{DC}}$ & \textbf{*0.709} & \textbf{*0.401} & \textbf{*0.663} \\
 & Q$_{\text{R}}$  & 0.696 & 0.386 & 0.649 \\
\bottomrule
\end{tabular}}
\caption{Cross-lingual zero-shot chunk retrieval performance (Non-English Queries to English Knowledge Base). Bold text indicates the best performance within each model group. * denotes the overall global maximum for each metric.}
\label{tab:retr_crosslingual}
\end{table}

\subsection{Generation experiment details} \label{appendix:generation_details}

For both the monolingual and cross-lingual setting we employ a subset of 400 dialogues (100 dialogues per language). As for the retrieval experiment, we employ as queries all hater's turns followed by a counterspeech grounded on external knowledge (we discard counterspeech examples not grounded on any external knowledge such as clarifying questions). Table \ref{tab:nr_queries_lang} and \ref{tab:nr_queries_target} shows the distribution of queries per language and per target, respectively.
We use the following hyperparameters for the Qwen3 8B model: max\_new\_tokens=100, temperature=0.7, top\_p=0.8, top\_k=20, min\_p = 0. Generations took roughly 1 hour to run on a NVIDIA A40, Ampere GPU.
Below we show the prompts we employed (for each prompt mentioning the dialogue history, this mention was not included for the generation of the first counterspeech where there was no dialogue history beyond the given HS):
\begin{itemize}
    \item \textbf{System Prompt}
    \begin{lstlisting}
Given the hateful claim provided in HS, and the dialogue history, generate a two-sentence-long CS in {language} following the guidelines provided in GUIDELINES.

GUIDELINES: Avoid abusive language and divisive labels (e.g., racist, fascist). Challenge the claim, not the person who wrote it. Counter misinformation with accurate facts, evidence, impartiality, and reliable sources. Provide context for the misinformed hateful claim. Express support for those under attack and respond with kindness and respect. Empathize with underlying fears or anxieties that caused the expression of hate. Challenge negative stereotypes using facts and providing context.
\end{lstlisting}

\item User prompt without knowledge (CS$_{base}$)
\begin{lstlisting}
Dialogue History: {dialogue_history}

Current HS: {hateful_message}

Task: Based on the current HS and dialogue history, generate the two-sentence counterstatement (CS) in {language}
\end{lstlisting}

\item User prompt with knowledge: knowledge used is gold for CS$_{gold}$ and the 5 top retrieved chunks for CS$_{retr}$
\begin{lstlisting}
Dialogue History: {dialogue_history}

GROUND: {knowledge}

Current HS: {hateful_message}

Task: Based on the current HS, dialogue history, and GROUND, generate the two-sentence counterstatement (CS) in {language}. The GROUND context consists of several distinct, isolated text chunks. You must necessarily use the facts contained in the GROUND chunks to contrast misinformation and stereotypes. Answer by referring exclusively to GROUND chunks, don't cite the sources in brackets.
\end{lstlisting}
\end{itemize}

\begin{table}[]
\small
\begin{tabular}{lcc}
\toprule
\textbf{Language}                           & \textbf{\# Q Mono.} & \textbf{\# Q Cross.} \\
\midrule
 English  & 243  & -      \\
                              Italian  & 235   & 230     \\
                             Polish   & 343  & 243       \\
                              Spanish  & 343    & 240      \\
                         Maltese  & -& 237        \\
                               \bottomrule
\end{tabular}
\caption{Distribution of language per queries in monolingual and cross-lingual setting for the generation experiment.}
\label{tab:nr_queries_lang}
\end{table}

\begin{table}[]
\small
\begin{tabular}{lcc}
\toprule
\textbf{Target} & \textbf{\# Q Mono.} & \textbf{\# Q Cross.} \\
\midrule
DISABLED        & 88                              & 147                               \\
JEWS            & 49                              & 45                                \\
LGBT+           & 243                             & 165                               \\
MIGRANTS        & 244                             & 137                               \\
MUSLIMS         & 163                             & 148                               \\
POC             & 63                              & 149                               \\
ROMANI          & 7                               & -                                 \\
WOMEN           & 214                             & 139                               \\
Multitarget     & 8                               & 20                               \\
\bottomrule
\end{tabular}
\caption{Distribution of target per queries in monolingual and cross-lingual setting for the generation experiment.}
\label{tab:nr_queries_target}
\end{table}

\paragraph{Evaluation metrics details} For \textbf{BERTScore} we report $F_1$ scores using \texttt{xlm-roberta-large}, for \textbf{NLI Entailment} we calculate the entailment of the CS by the preceding HS with \texttt{xlm-roberta-large-xnli} \cite{joeddav2020xlmrobertaxnli}. We employ \texttt{gpt-4.1-mini} as \textbf{LLM-as-a-judge} to measure Faithfulness and Answer relevance, with max\_output\_tokens=20 and temperature=0. We employ an adapted version of the definitions given by \citet{es2024ragas} in the prompts to calculate these metrics, but we add a score rubric to more clearly direct model scoring. In particular, for \textbf{Faithfulness} we employ the following prompts:
\begin{itemize}
\item System Prompt:
\begin{lstlisting}
Faithfulness measures the information consistency of the answer against the given context. Any claims that are made in the answer that cannot be deduced from context should be penalized. Given an answer and context, assign a score for faithfulness in the range 1-5.

### Score Rubrics [Faithfulness]
Score 1: The answer contains major contradictions, fabricated information, or unsupported claims. Most of the content cannot be verified from the context.
Score 2: The answer includes several unsupported or inaccurate claims. Important details are invented, exaggerated, or inconsistent with the context.
Score 3: The answer is mostly grounded in the context, but includes some unsupported assumptions, minor hallucinations, or overgeneralizations.
Score 4: The answer is highly consistent with the context and contains only small ambiguities or negligible unsupported additions that do not materially affect accuracy.
Score 5: Every claim in the answer is directly supported by or can be clearly inferred from the context. The answer contains no hallucinations, contradictions, or misleading interpretations.

Return ONLY a single digit (1-5). No reasoning. No extra text.
\end{lstlisting}

\item User Prompt:
\begin{lstlisting}
Context: {ground}

Answer: {response}

Score: 
\end{lstlisting}
\end{itemize}

For \textbf{Answer relevance} we employ these prompts:

\begin{itemize}
\item System Prompt:
\begin{lstlisting}
Answer Relevancy measures the degree to which a counterstatement (CS) directly addresses and is appropriate for a given harmful statement (HS). It penalizes the presences of redundant information or incomplete CS given an HS. Given an HS and CS, assign a score for answer relevancy in the range 1-5.

### Score Rubrics [Answer Relevancy]
Score 1: The CS fails to address the HS or is mostly irrelevant. It may omit the main point entirely or provide unrelated information.
Score 2: The CS partially addresses the HS but includes substantial irrelevant, redundant, or distracting content. Important aspects of the HS are left unanswered.
Score 3: The CS addresses the main point of the HS but may be incomplete, somewhat unfocused, or contain unnecessary information that reduces clarity.
Score 4: The CS directly addresses the HS and is mostly complete. Minor redundancy or small omissions may be present, but the response remains focused and appropriate.
Score 5: The CS fully, directly, and efficiently addresses the HS. All relevant aspects are covered with no unnecessary, redundant, or off-topic information.

Return ONLY a single digit (1-5). No reasoning. No extra text.
\end{lstlisting}

\item User Prompt:
\begin{lstlisting}
HS: {hs}

CS: {response}

Score:
\end{lstlisting}
\end{itemize}

\paragraph{Generation Cross-lingual results} Table \ref{tab:gen_results_cross} presents the cross-lingual results for the generation experiment.

\begin{table}[ht]
\small
\centering
\begin{tabular}{llr}
\toprule
\textbf{Metric} & \textbf{Setting} & \textbf{Value} \\
\midrule
                  \multirow{3}{*}{BERTscore}  &          CS$_{gold}$ &  \textbf{0.882} \\
                  &                   CS$_{base}$ &  0.870 \\
                   &          CS$_{retr}$ &  0.874 \\
                   \midrule
               \multirow{3}{*}{Faithfulness$_{gold}$} &          CS$_{gold}$ &  \textbf{3.486} \\
                &                   CS$_{base}$ &  2.183 \\
                &          CS$_{retr}$ &  2.507 \\
                \midrule
 Faithfulness$_{retr}$ &          CS$_{retr}$ &  \textbf{3.858} \\
 \midrule
             \multirow{3}{*}{NLI Entailment$_{gold}$} &          CS$_{gold}$ &  \textbf{0.202} \\
 &                   CS$_{base}$ &  0.045 \\
             &          CS$_{retr}$ &  0.077 \\
             \midrule
NLI Entailment$_{retr}$ &          CS$_{retr}$ &  \textbf{0.212} \\
\midrule
                  \multirow{3}{*}{Relevance} &          CS$_{gold}$ &  3.855 \\
                   &                   CS$_{base}$ & \textbf{3.917} \\
                   &          CS$_{retr}$ &  3.706 \\
\bottomrule
\end{tabular}
\caption{CS generation quality across metrics (cross-lingual). Bold values represent the maximum score achieved within each evaluation category.}
\label{tab:gen_results_cross}
\end{table}

\end{document}